%% file: EGauthorGuidelines-cgf-sub.tex
\documentclass{egpubl}
 
\JournalSubmission    % uncomment for submission to Computer Graphics Forum
\usepackage[T1]{fontenc}
\usepackage{dfadobe}  

\usepackage{cite}  % comment out for biblatex with backend=biber
\BibtexOrBiblatex
\electronicVersion
\PrintedOrElectronic
\ifpdf \usepackage[pdftex]{graphicx} \pdfcompresslevel=9
\else \usepackage[dvips]{graphicx} \fi

\usepackage{egweblnk} 
\title[A Survey on 3D Human Avatar Modeling - From Reconstruction to Generation]%
      {A Survey on 3D Human Avatar Modeling - From Reconstruction to Generation}

\author[R. Wang \& Y. Cao et al]{
Ruihe Wang$^{2*}$ 
Yukang Cao$^{1*}$\thanks{Corresponding author. $^{*}$ Equal contributions.}
Kai Han$^{1}$
Kwan-Yee K. Wong$^{1}$
\\ $^{1}$ The University of Hong Kong $^{2}$ Harbin Institute of Technology}
\usepackage{amsmath}
\DeclareMathOperator*{\argmax}{arg\,max}
\usepackage{amsfonts}
\usepackage{booktabs}
\usepackage{multirow}

\newcommand{\fref}[1]{Fig.~\ref{#1}}
\newcommand{\tref}[1]{Tab.~\ref{#1}}
\newcommand{\Sref}[1]{Sec.~\ref{#1}}
\newcommand{\sref}[1]{Sec.~\ref{#1}}
\newcommand{\Eref}[1]{Equation (\ref{#1})}

\begin{document}

% uncomment for using teaser
% \teaser{
%  \includegraphics[width=0.9\linewidth]{eg_new}
%  \centering
%   \caption{New EG Logo}
% \label{fig:teaser}
%}

\maketitle
%-------------------------------------------------------------------------
\begin{abstract}
   3D modeling has long been an important area in computer vision and computer graphics. Recently, thanks to the breakthroughs in neural representations and generative models, we witnessed a rapid development of 3D modeling. 3D human modeling, lying at the core of many real-world applications, such as gaming and animation, has attracted significant attention. Over the past few years, a large body of work on creating 3D human avatars has been introduced, forming a new and abundant knowledge base for 3D human modeling. The scale of the literature makes it difficult for individuals to keep track of all the works. This survey aims to provide a comprehensive overview of these emerging techniques for 3D human avatar modeling, from both reconstruction and generation perspectives. Firstly, we review representative methods for 3D human reconstruction, including methods based on pixel-aligned implicit function, neural radiance field, and 3D Gaussian Splatting, etc. We then summarize representative methods for 3D human generation, especially those using large language models like CLIP, diffusion models, and various 3D representations, which demonstrate state-of-the-art performance. Finally, we discuss our reflection on existing methods and open challenges for 3D human avatar modeling, shedding light on future research.
%-------------------------------------------------------------------------
%  ACM CCS 1998
%  (see https://www.acm.org/publications/computing-classification-system/1998)
% \begin{classification} % according to https://www.acm.org/publications/computing-classification-system/1998
% \CCScat{Computer Graphics}{I.3.3}{Picture/Image Generation}{Line and curve generation}
% \end{classification}
%-------------------------------------------------------------------------
%  ACM CCS 2012
   % (see https://www.acm.org/publications/class-2012)
%The tool at \url{http://dl.acm.org/ccs.cfm} can be used to generate
% CCS codes.
%Example:
% \begin{CCSXML}
% <ccs2012>
% <concept>
% <concept_id>10010147.10010371.10010352.10010381</concept_id>
% <concept_desc>Computing methodologies~Collision detection</concept_desc>
% <concept_significance>300</concept_significance>
% </concept>
% <concept>
% <concept_id>10010583.10010588.10010559</concept_id>
% <concept_desc>Hardware~Sensors and actuators</concept_desc>
% <concept_significance>300</concept_significance>
% </concept>
% <concept>
% <concept_id>10010583.10010584.10010587</concept_id>
% <concept_desc>Hardware~PCB design and layout</concept_desc>
% <concept_significance>100</concept_significance>
% </concept>
% </ccs2012>
% \end{CCSXML}

% \ccsdesc[500]{Computing methodologies~Computer vision}
% % \ccsdesc[300]{Computing methodologies~Artificial intelligence}
% % \ccsdesc[100]{Computing methodologies~Computer graphics}

% % \ccsdesc[300]{Hardware~Sensors and actuators}
% % \ccsdesc[100]{Hardware~PCB design and layout}

% \printccsdesc   
\end{abstract}
%-------------------------------------------------------------------------
\input{Sections/1_introduction}

\input{Sections/scope}

\input{Sections/2_Implicit_function-based/Implicit_function-based}

\input{Sections/3_NeRF-based/NeRF-based}

\input{Sections/4_Gaussian/Gaussian}

\input{Sections/5_GAN-based/GAN-based}

\input{Sections/6_CLIP-based/CLIP-based}

\input{Sections/7_Diffusion_model-based/Diffusion_model-based}

\input{Sections/8_Reflection/Reflection}

\bibliographystyle{eg-alpha-doi} 
\bibliography{reb}

\end{document}

%% file: Sections/1_introduction.tex
\section{Introduction}\label{sec:introduction}

Human avatar modeling has recently shown significant scientific progress, with diverse applications ranging from computer graphics and gaming to virtual reality and medical imaging. While early methods rely on expensive capturing hardware and labor-intensive calibration processes to produce good-looking models~\cite{collet2015high}, recent advancements have made it much more convenient to reconstruct and generate human avatars from various types of input such as images, videos, or text prompts.

\begin{figure*}[t]
  \centering
  \includegraphics[width=\textwidth]{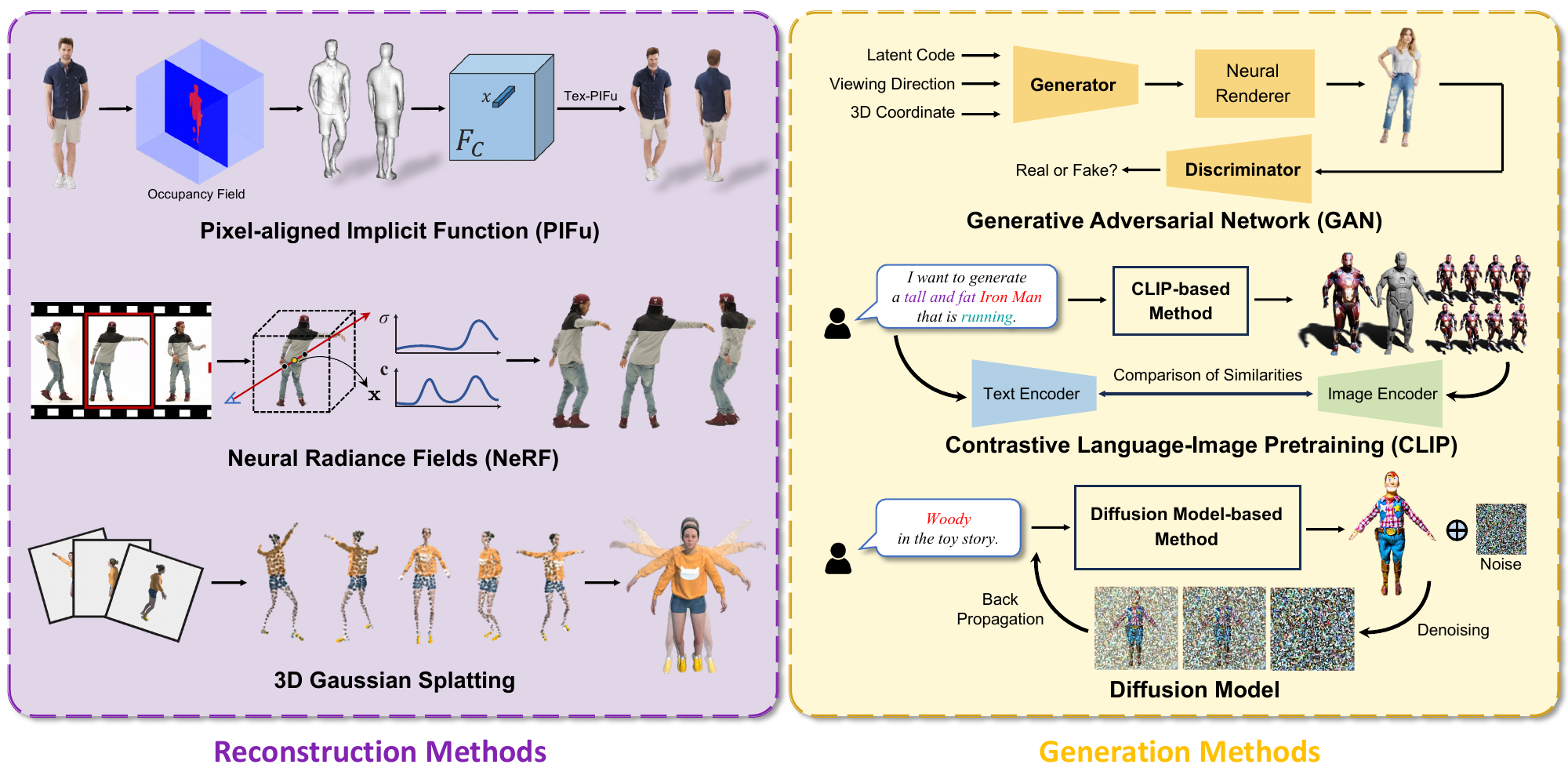}
  \caption{An overview of typical 3D human avatar modeling approaches. Example images adapted from~\cite{saito2019pifu, weng2022humannerf, huang2023sc,zhang2022avatargen, hong2022avatarclip, cao2023dreamavatar}.}
  \label{fig:overview-survey}
\end{figure*}

3D human mesh reconstruction methods can roughly be categorized into model-based methods~\cite{anguelov2005scape, balan2007detailed, bhatnagar2019multi, alldieck2019tex2shape} and model-free methods~\cite{chen2019learning, chibane2020implicit, deng2020nasa, peng2020convolutional}. Model-based methods involve fitting an explicit parametric human model (e.g., SMPL~\cite{SMPL:2015}) to an image but they encounter challenges in capturing intricate details like clothing and hair. In contrast, model-free methods overcome these limitations by predicting the occupancy values of a volumetric space.
A representative method is PIFu~\cite{saito2019pifu}, which exploits a Multi-Layer Perceptron (MLP) to model an implicit function that predicts the occupancy value of a given point by leveraging pixel-aligned features extracted from the input image. However, PIFu does not leverage the human body’s structure, thus struggling with challenging poses, self-occlusions, and depth ambiguities. Follow-up works~\cite{saito2020pifuhd, he2020geopifu, hong2021stereopifu, zheng2021pamir, Huang_2020_CVPR, he2021arch++, cao2022jiff, xiu2022icon, cao2023sesdf} address these shortcomings by integrating priors such as normal maps, SMPL model, and depth information. However, these methods still face a topological constraint that restricts the model's performance with loose clothing, which is subsequently addressed by ECON~\cite{xiu2023econ} via an explicit method. Besides methods utilizing a single-view image as input, multi-view scenarios~\cite{zheng2021deepmulticap, shao2022diffustereo} offer a richer source of information from different viewpoints, leading to improved reconstruction results.

However, the performance of PIFu-based methods is largely limited by the quality of 3D training datasets, which are scarce and difficult to obtain. NeRF~\cite{mildenhall2021nerf} enables novel-view synthesis by simply inputting a limited set of images to obtain the RGB color and density value of each 3D point. Building upon NeRF, researchers proposed numerous 3D human reconstruction methods~\cite{weng2022humannerf, peng2021neural, kwon2021neural, liu2023hosnerf} that represent a 3D human as a neural radiance field without relying on prior knowledge or pre-trained models. Beyond reconstruction, the exploration of synthesizing free-viewpoint animations with user-controlled novel pose sequences is also a promising research direction~\cite{liu2021neural, peng2021animatable, li2022tava, 2021narf, wang2022arah, weng2022humannerf}. Furthermore, methods~\cite{suo2021neuralhumanfvv, tao2021function4d, shao2022doublefield} have also been proposed to integrate both surface and radiance fields to achieve high-fidelity 3D human novel view synthesis.

Yet, achieving high-quality reconstruction results using neural radiance fields still demands neural networks that are costly to train and render. 3D Gaussian Splatting (3DGS)~\cite{kerbl20233d} represents and renders complex scenes in a shorter training time without sacrificing speed for quality. By modeling a scene via a set of 3D Gaussians, Kerbl \emph{et al.}~\cite{kerbl20233d} employ an explicit and object-centric method that is different from implicit representations like NeRF and \textsc{DMTet}~\cite{shen2021deep}. Following 3DGS, many works~\cite{liu2023animatable, zielonka2023drivable, li2023human101, xu2023gaussian, huang2023sc, hu2023gauhuman} have been proposed, utilizing its core principles for enhanced 3D human reconstruction, leading to highly animatable and realistic human models.

The advent of Generative Adversarial Networks (GANs)~\cite{goodfellow2014generative} marked an era of human avatar generation. GAN-based generation methods~\cite{bergman2022generative, zhang2022avatargen, noguchi2022unsupervised, jiang2022humangen, hong2022eva3d} are typically composed of two key components: the StyleGAN~\cite{fu2022stylegan} architecture and triplane representation proposed in EG3D~\cite{chan2022eg3d}, successfully establishing a connection between 3D fields and 2D images. As a result, these methods have achieved remarkable progress in the field of 3D human avatar generation, despite being trained solely on 2D datasets.

While GAN-based methods have yielded impressive results, they still lack the ability to generate unseen characteristics that are not contained in the training dataset. With the recent development of large-language models, 3D generative methods~\cite{jain2022dreamfeild, michel2022text2mesh, jiang2022text2human, hong2022avatarclip} employ CLIP~\cite{radford2021learning} to generate 3D contents directly from text prompts. However, due to CLIP's limitation in fully comprehending textual descriptions, these methods still struggle to generate humans with fine details and complex motions. The advent of diffusion models has significantly advanced 3D generation by converting Gaussian noise into structured data via a Markov process with denoising steps. Drawing inspiration from DreamFusion's Score Distillation Sampling (SDS) technique~\cite{poole2022dreamfusion}, many methods~\cite{cao2023dreamavatar, jiang2023avatarcraft, huang2023dreamwaltz, kolotouros2023dreamhuman, weng2023zeroavatar, liao2023tada, zeng2023avatarbooth, huang2023humannorm} are proposed to enhance generation quality from various perspectives, driving advancements of this field. Meanwhile, researchers have also introduced additional possibilities that extend the use of diffusion models to facilitate controllable 3D human editing~\cite{shao2023control4d, han2023headsculpt, liu2023headartist, pan2023avatarstudio}.

\begin{figure*}[h]
  \centering
  \includegraphics[width=\textwidth]{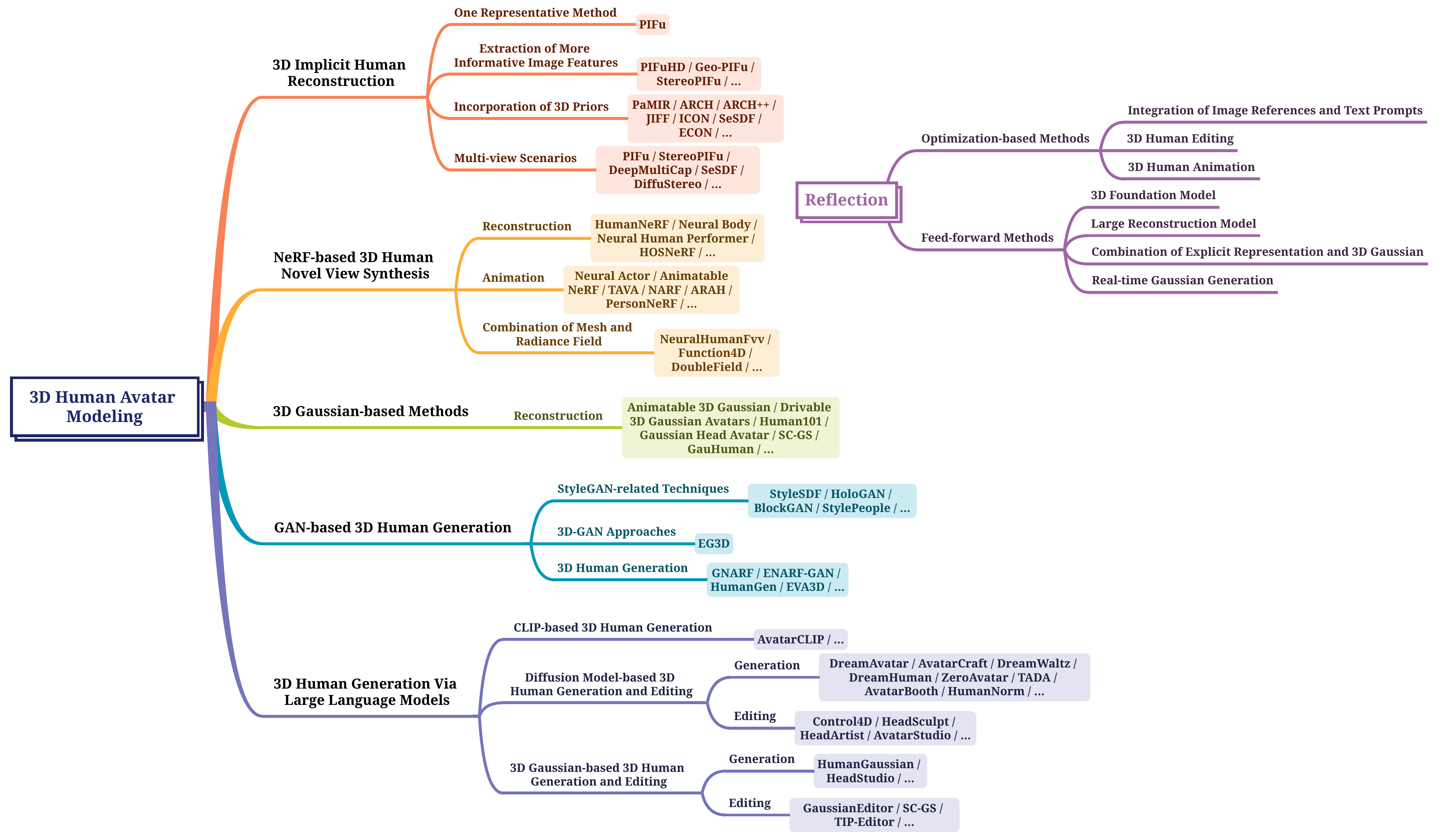}
  \caption{Taxonomy of 3D human avatar modeling methods in this survey.}
  \label{fig:overview_survey}
\end{figure*}

We provide an overview of 3D human modeling via various 3D representations in~\fref{fig:overview-survey}. In this paper, we present a taxonomy of recent research that maps out the evolution process of 3D human avatar modeling, categorizing it into five key areas: PIFu-based 3D implicit human reconstruction (\Sref{sec: pifu}), NeRF-based 3D human novel view synthesis (\Sref{sec:nerf}), 3D Gaussian-based methods (\Sref{sec:gaussian}), GAN-based 3D human generation (\Sref{sec:gan}), and language model-based 3D human generation and editing (\Sref{sec:generation}). We further examine the models by breaking each category into more detailed sub-categories (see~\fref{fig:overview_survey} for an overview of this survey). In \Sref{sec:reflection}, we reflect upon existing methods and discuss the open challenges and potential directions for future research, focusing primarily on two categories of approaches: optimization-based methods and feed-forward approaches. 

In summary, this paper offers a comprehensive survey for 3D human avatar modeling. 
Particularly, we make the following contributions:
Firstly, we present a thorough and up-to-date review of representative methods for 3D human reconstruction. 
Secondly, we summarize representative methods for the emerging techniques for 3D human generation. 
Finally, we provide our reflections on existing methods for 3D human avatar modeling and discuss insights and potential future research directions for future development of this field.

%% file: Sections/scope.tex
\section{Scope of This Survey}

This survey delves into recent advancements in 3D human modeling that utilize neural networks. Specifically, we start with illustrating the process of implicit function-based 3D human reconstruction from monocular images. Subsequently, we analyze the impact of the neural radiance field (NeRF) and 3D Gaussian Splatting on 3D human modeling. We then explore the realm of 3D generative AI, with a specific focus on the generative adversarial network (GAN), contrastive language-image pretraining (CLIP), and diffusion models. At the end of this survey, we provide our insights into future directions in this field. 

This survey comprehensively discusses the essential techniques in 3D human modeling, covering 3D reconstruction, generation, and editing, to provide a detailed understanding of the past, present, and future. We have collected papers from major computer vision and computer graphics conferences and journals, as well as preprints available on arXiv. The selection process prioritized relevance to the scope of this survey, aiming to offer an inclusive overview of the rapid advances in this field. However, it is important to note that while this report serves as a compilation of state-of-the-art methods in a specific domain, it is hard to have a complete coverage due to the vast number of publications and the rapid development of the field. Readers are encouraged to refer to the cited works for more in-depth discussions and additional methodologies.

\noindent \textbf{Related Survey.}
Human mesh recovery (HMR) is fundamental to current 3D human modeling. However, due to space limit, we consider them to be beyond the scope of this report. Interested readers can find more comprehensive insights on HMR in \cite{wang2021deep, zheng2023deep, tian2023recovering}. 3D representations form the foundational basis for both 3D human modeling and the construction of 3D general objects. We encourage readers to learn from \cite{guo2020deep, huang2021comprehensive} for the details of point clouds, \cite{tewari2020state, tewari2022advances} for NeRF, and \cite{chen2024survey, fei20243d, wu2024recent} for the most recent 3D Gaussian Splatting. Furthermore, the advancement of large language models has spurred research in another facet of the 3D virtual realm, namely, 3D general objects. Recent progress in 3D generative approaches can be explored through \cite{yang2023diffusion, cao2024survey, li2023generative, li2024advances}. We also recommend consulting \cite{croitoru2023diffusion, xing2023survey, po2023state} for insights into 2D generative AI.

%% file: Sections/2_Implicit_function-based/Implicit_function-based.tex
\section{3D implicit human reconstruction}
\label{sec:pifu}

In this section, we first discuss methods for 3D human reconstruction based on implicit function, which can accurately capture clothing topologies from single or multiple images. We start with pixel-aligned implicit function (PIFu), which is one of the most representative works in this regime, in \sref{sec: pifu}. We then explore (1) enhancements at the feature level in \sref{sec: pifu-image-feature-level}, (2) the usage of 3D priors in \sref{sec: pifu-3d-prior}, and (3) multi-view scenarios in \sref{sec: pifu-multiview}.

\subsection{Pixel-aligned Implicit Function}
\label{sec: pifu}

\begin{figure}[t]
  \centering
  \includegraphics[width=8.5cm]{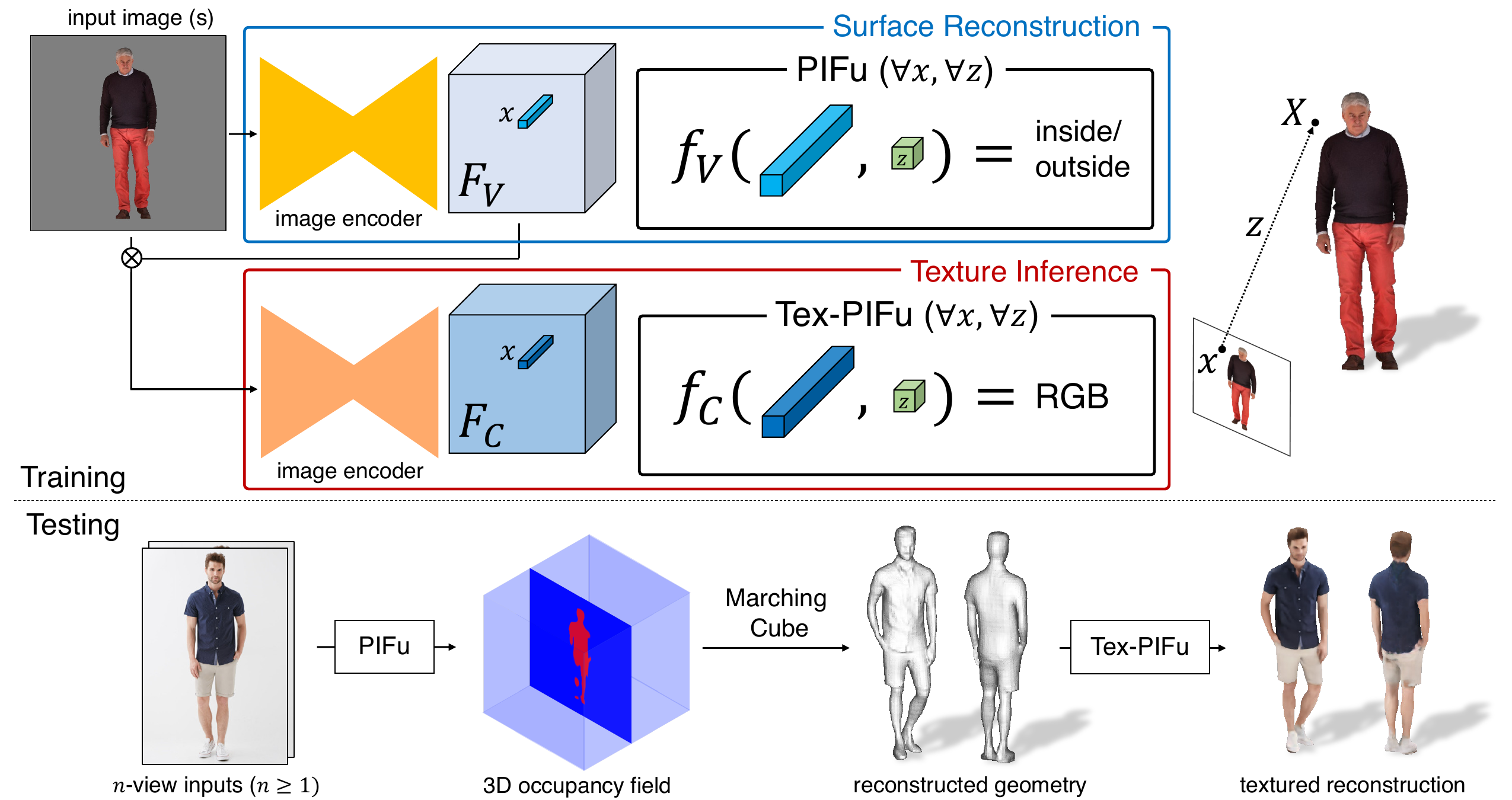}\\
  \caption{An overview of PIFu network for predicting surface and texture values. Figure obtained from~\cite{saito2019pifu}.}
  \label{fig:pifu}
\end{figure}

Given a single image or multiple images as input, PIFu~\cite{saito2019pifu} applies an hour-glassed image encoder~\cite{newell2016stacked} and bi-linear interpolation to extract pixel-aligned image features for each 3D query point $\mathbf{x} \in \mathbb{R}^3$ and then predicts a continuous 3D occupancy field via an implicit function (see \fref{fig:pifu}). Specifically, the implicit function, implemented as an MLP, defines the 3D human surface as the level set of the function, e.g., $F(\mathbf{x}) = 0.5$. It can be written as:
\begin{equation}
    F_{V}(\mathcal{B}(\psi_{\text{geo}}(I), \pi(\mathbf{x})), z(\mathbf{x})) \mapsto s: s \in \mathbb{R},
\end{equation}
where $\mathcal{B}(\psi_{\text{geo}}(I), \pi(\mathbf{x}))$ is the pixel-aligned feature, with $\psi_{\text{geo}}(I) \in \mathbb{R}^{128 \times 128 \times 256}$ denoting the feature map extracted from the image $I \in \mathbb{R}^{512 \times 512 \times 3}$. $\pi(\mathbf{x})$ is the 2D projection of $\mathbf{x}$ on the image, and $z(\mathbf{x})$ is the depth value.

After obtaining the geometry in form of a mesh by applying Marching Cube~\cite{lorensen1998marching} to the occupancy field, PIFu can be extended to texture inference by predicting the RGB value $\mathbf{c}$ for each 3D vertices:
\begin{equation}
    F_{C}(\mathcal{B}(\psi_{\text{tex}}(I, f_{\text{geo}}), \pi(\mathbf{x})), z(\mathbf{x})) \mapsto \mathbf{c}: \mathbf{c} \in \mathbb{R}^3,
\end{equation}
where $f_{\text{geo}}$ is the geometry feature obtained from the geometry inference stage.

Besides single-view 3D human reconstruction, PIFu can also take calibrated multi-view images as inputs (suppose we have $n$-view images as inputs). For each 3D query point $\mathbf{x}$, we first obtain the image features $\Phi_{\text{view}_{i}} = \mathcal{B}(\psi(I_{i}), \pi_i(\mathbf{x}))$ from the $i$-th view. The average feature $\mathtt{avg}\{\cdot\}$ across all the views is utilized to enhance prediction accuracy for both surface and texture:
\begin{equation}
F(\mathtt{avg}\{\Phi_{\text{view}_{1}},\ldots,\Phi_{\text{view}_{n}}\})\mapsto s.
\end{equation}

While PIFu shows potential in 3D human reconstruction via implicit functions, it encounters two major challenges: (1) The reconstructed 3D mesh still shows sub-optimal geometry details for real-world applications. (2) Due to the lack of 3D information under single-view scenarios, PIFu falls short in handling complex poses and self-occlusions, and suffers from depth ambiguity issues. Subsequent methods aim to overcome these limitations by extracting more informative image features and incorporating 3D priors.

\subsection{Extraction of More Informative Image Features}
\label{sec: pifu-image-feature-level}

\textbf{PIFuHD}
To enhance the quality of clothing topologies, PIFuHD~\cite{saito2020pifuhd} introduces a coarse-to-fine network that integrates both low- and high-resolution image feature maps as well as the predicted front- and back-side normal images. Different from PIFu which takes low-resolution images $I_L \in \mathbb{R}^{512 \times 512 \times 3}$ as input, PIFuHD opts for information from high-resolution images, i.e., $I_H \in \mathbb{R}^{1024 \times 1024 \times 3}$. At the coarse stage, PIFuHD modifies PIFu by concatenating the predicted front and back normal maps ($\mathbf{n}_{L, \text{front}}$ and $\mathbf{n}_{L, \text{back}}$) with the downsampled low-resolution image $I_L$ as inputs:
\begin{equation}
    F_L(\mathcal{B}(\psi(I_L \oplus \mathbf{n}_{L, \text{front}} \oplus \mathbf{n}_{L, \text{back}}), \pi(\mathbf{x})), z(\mathbf{x})) \mapsto s_L,
\end{equation}
where $\oplus$ denotes the concatenation operation.
PIFuHD then operates over the original high-resolution input at the fine stage:
\begin{equation}
    F_H(\mathcal{B}(\psi(I_H \oplus \mathbf{n}_{H, \text{front}} \oplus \mathbf{n}_{H, \text{back}}), \pi(\mathbf{x})), \Omega(\mathbf{x})) \mapsto s_H,
\end{equation}
where $\Omega(X)$ denotes the intermediate output from the penultimate layer of the MLP in the coarse branch.

\textbf{Geo-PIFu}
Geo-PIFu~\cite{he2020geopifu} proposes to extract both 3D space-aligned and 2D pixel-aligned features from the input image, effectively encoding the structural information for occupancy estimation:
\begin{equation}
    F(\mathcal{T}(\xi(I),\mathbf{x}),\mathcal{B}(\psi(I), \pi(\mathbf{x})), z(\mathbf{x})))\mapsto s,
\end{equation}
where $\xi$ is the 3D U-Net~\cite{jackson2017large}, with $\mathcal{T}(\cdot)$ denoting the multi-scale tri-linear interpolation. However, Geo-PIFu still struggles with capturing high-resolution clothing details since the information employed is extracted only from the image.

\textbf{StereoPIFu}
StereoPIFu~\cite{hong2021stereopifu} targets at reconstructing 3D implicit human models from binocular images. Specifically, given the multi-view stereo inputs, StereoPIFu proposes a stereo vision-based network to extract voxel-aligned features. This geometric prior, together with a novel relative z-offset and depth maps, are utilized to produce enhanced reconstruction results:
\begin{equation}
    F(\mathcal{B}(\psi_l(I), \pi_{l}(\mathbf{x})),\Phi(\mathbf{x}),\Psi(\mathbf{x}),\eta(Z_{E}(\mathbf{x}))) \mapsto s,
\end{equation}
where $\mathcal{B}(\psi_l(I), \pi_{l}(\mathbf{x}))$ denotes pixel-aligned features from the left image, $\Phi(\mathbf{x})$ and $\Psi(\mathbf{x})$ are voxel-aligned features, and $\eta(\cdot)$ is a transformation function designed to normalize the relative z-offset $Z_{E}$ to the interval $(-1.0, 1.0)$::
\begin{equation}
    Z_{E} = z(\mathbf{x}) - E(\pi_{l}(\mathbf{x})),
\end{equation}
where $E(\cdot)$ is the predicted depth map of the left image.

\subsection{Incorporation of 3D Priors}
\label{sec: pifu-3d-prior}

Attempts have also been made to incorporate 3D features as priors to enhance robustness, with SMPL~\cite{SMPL:2015} and SMPL-X~\cite{SMPL-X:2019} being frequently employed.
Specifically, given the pose parameter $\theta$ and shape parameter $\beta$, SMPL can map the canonical model with $n_{S}$ vertices to observation space\footnote{In this survey, we refer to the space with poses of the SMPL model that differ from canonical pose as the observation space.}:
\begin{subequations} \label{smpl}
    \begin{align}
        M(\beta,\theta)&=\mathtt{lbs}(T(\beta,\theta),J(\beta), \theta,\mathcal{W}),\\
        T(\beta,\theta)&=\mathbf{T}+B_{s}(\beta)+B_{p}(\theta),
    \end{align}
\end{subequations}
where $M$ is the function representing the SMPL model in the observation space, and $T$ gives the transformed vertices.
$\mathcal{W}$ is the blend weight, $B_s$ and $B_p$ are the shape blend shape function and pose blend shape function, respectively. $\mathtt{lbs}(\cdot)$ denotes the linear blend skinning function, corresponding to articulated deformation. It poses $T(\cdot)$ based on the pose parameters $\theta$ and joint locations $J(\beta)$, using the blend weights $\mathcal{W}$, individually for each body vertex:
\begin{equation}\label{eq:deformation}
    % \small
    \mathbf{v}_o = \mathcal{G} \cdot \mathbf{v}_c, \quad \mathcal{G} = \sum_{k=1}^{K} w_k \mathcal{G}_k (\theta, {j}_k),
\end{equation}
where $\mathbf{v}_c$ and $\mathbf{v}_o$ respectively are SMPL vertices under the canonical pose and observation space, $w_k$ is the skinning weight, $\mathcal{G}_k (\theta, j_k)$ is the affine deformation that transforms the $k$-th joint ${j}_k$ from the canonical space to observation space, and $K$ is the number of neighboring joints.

SMPL-X evolves from SMPL to include more face vertices, expression parameters $\phi$ and the expression blend shape function $B_e$ into the model:
\begin{subequations} \label{smplx}
    \begin{align}
        M(\beta, \theta, \phi)&=\mathtt{lbs}(T(\beta, \theta, \phi), J(\beta), \theta, \mathcal{W}), \\ 
    T(\beta, \theta, \phi)&=\mathbf{T} + B_s(\beta) + B_e(\phi) + B_p(\theta).
    \end{align}
\end{subequations}

\textbf{PaMIR}
PaMIR~\cite{zheng2021pamir} is one of the first to extract 3D voxel-aligned features and semantic information from SMPL to provide a 3D-aware prior and address complex poses and self-occlusion issues.
Specifically, PaMIR first estimates an initial SMPL model from the input image $I$ via a pre-trained GCMR network~\cite{kolotouros2019convolutional}. For each 3D query point $\mathbf{x}$, its occupancy value is then predicted based on both pixel-aligned and voxel-aligned features:
\begin{equation}
    F(\mathcal{B}(\psi(I), \pi(\mathbf{x})), \mathcal{T}(f_{3D}, \mathbf{x})) \mapsto s,
\end{equation}
where $\mathcal{T}(f_{3D}, \mathbf{x})$ represents the voxel-aligned feature. The 3D feature volume $f_{3D}$ is derived by subsequently (1) converting the SMPL mesh into an occupancy volume $V_s$ through mesh voxelization; (2) encoding the occupancy volume through a 3D encoder $E_{3D}$, i.e., $f_{3D}=E_{3D}(V_s)$. However, PaMIR still performs poorly in reconstructing dynamic and complex movements when hands and clothes are close to each other.

\textbf{ARCH}
ARCH~\cite{Huang_2020_CVPR}, on the other hand, utilizes the Semantic Space (SemS) and the Semantic Deformation Field (SemDF) to transform query points from the observation space to the canonical space before calculating the occupancy value. It also propose to extract the spatial feature based on SemS via Radial Basis Function (RBF).
An occupancy sub-network $F_s$, a normal sub-network $F_\mathbf{n}$, and a color sub-network $F_\mathbf{c}$ are then utilized for implicit surface reconstruction in the canonical space:
\begin{subequations}
    \begin{gather}
        F_s(\mathcal{B}(f_{2D}, \pi(\mathbf{x})), \mathcal{T}(f_{3D}, \mathbf{x})) \mapsto s, \\
        F_\mathbf{n}(\mathcal{B}(f_{2D}, \pi(\mathbf{x})), \mathcal{T}(f_{3D}, \mathbf{x}), f_s) \mapsto \mathbf{n}, \\
        F_\mathbf{c}(\mathcal{B}(f_{2D}, \pi(\mathbf{x})), \mathcal{T}(f_{3D}, \mathbf{x}), f_s, f_\mathbf{n}) \mapsto \mathbf{c},
    \end{gather}
\end{subequations}
where $f_{2D}$ is the 2D feature map, $f_{3D}$ is the 3D feature volume, $f_s$ and $f_\mathbf{n}$ are feature maps extracted from occupancy and normal sub-networks.
Despite adding more information from various poses, ARCH can only reconstruct 3D humans under canonical space. Warping the mesh back to the observation space often results in artifacts such as intersecting surfaces and distorted body parts, leading to the degeneration of the reconstruction fidelity.

\textbf{ARCH++}
Adopting the deformation field from ARCH, ARCH++~\cite{he2021arch++} proposes to learn the joint-space occupancy field in both observation and canonical spaces. It first transforms the posed mesh to the canonical space, and then uniformly samples query points on the mesh surface. For each query point $\mathbf{x}_c$, ARCH++ applies tri-linear interpolation to obtain spatial-aligned features $f_{3D}$, which are encoded by the PointNet++~\cite{qi2017pointnet, qi2017pointnet++}.
The occupancy value is then jointly predicted in both the observation and canonical spaces to obtain additional constraints on the cross-space consistency:
\begin{subequations}
    \begin{align}
        F_o\left(\mathcal{B}(f_{2D}, \pi(\mathbf{x})), \mathcal{T}(f_{3D}, \mathbf{x})\right)&\mapsto s_o, \\
        F_c\left(\mathcal{B}(f_{2D}, \pi(\mathbf{x})), \mathcal{T}(f_{3D}, \mathbf{x})\right)&\mapsto s_c,
    \end{align}
\end{subequations}
where $s_o$, and $s_c$ stand for the occupancy value respectively in the observation and canonical space.

\textbf{JIFF}
JIFF~\cite{cao2022jiff} leverages the 3DMM~\cite{blanz19993dmm} as a 3D face prior to extract space-aligned 3D features with detailed geometry and texture information for improving the face quality. Given an input image $I$, JIFF first crops the face region and fits a 3DMM mesh $\mathbf{S}$ based on the cropped image. An encoder, which takes the vertices of 3DMM as input, is then employed to generate the 3D feature volume $\varphi(\mathbf{S})$. Thus, JIFF employs both pixel-aligned 2D features and space-aligned 3D features to complete the reconstruction:
\begin{equation}
    F(\mathcal{T}(\varphi(\mathbf{S}), \mathbf{x}), \mathcal{B}(\psi(I), \pi(\mathbf{x})), z(\mathbf{x})) \mapsto s.
\end{equation}

\textbf{ICON}
Despite incorporating 3D-aware prior into the implicit function, previous methods still fall short in processing complex poses.
To this end, ICON~\cite{xiu2022icon} replaces the global encoder with a more data-efficient local scheme. Specifically, given an image $I$ as input, ICON first uses a normal network $\mathcal{G}=(\mathcal{G}_\text{front}, \mathcal{G}_\text{back})$ to predict clothed body normal maps $\widehat{\mathcal{N}}=\{\widehat{\mathcal{N}}_{\text {front}}, \widehat{\mathcal{N}}_{\text {back}}\}$ based on $I$ and SMPL's front and back normal renderings:
\begin{equation}
    \mathcal{G}\left(\mathcal{R}_{\textbf{n}}(\text{SMPL}), I \right) \rightarrow \widehat{\mathcal{N}},
\end{equation}
where $\mathcal{R}_{\textbf{n}}(\cdot)$ is a PyTorch3D~\cite{ravi2020accelerating} differentiable renderer. An implicit representation of the surface is then regressed based on these local features:
\begin{equation}
    F(d(\mathbf{x}), \mathbf{n}(\mathbf{x}), \mathcal{N}(\pi(\mathbf{x}))) \mapsto s,
\end{equation}
in which $d(\cdot)$ is the signed distance from a point $\mathbf{x}$ to closest vertex $\mathbf{x}'$ of SMPL, $\mathbf{n}(\cdot)$ is the barycentric surface normal of  $\mathbf{x}'$, and $\mathcal{N}(\cdot)$ is a normal vector extracted from $\widehat{\mathcal{N}}^{\mathrm{c}}$ depending on the visibility of $\mathbf{x}'$:
\begin{equation}
    \mathcal{N}(\pi(\mathbf{x}))= \begin{cases}\mathcal{N}_{\text{front }}(\pi(\mathbf{x})), & \text{ if } \mathbf{x}' \text { is visible}, \\ \mathcal{N}_{\text {back }}(\pi(\mathbf{x})), & \text { otherwise}.\end{cases}
\end{equation}

\textbf{SeSDF}
SeSDF~\cite{cao2023sesdf} aims to flexibly and robustly extract detailed information of clothed humans from either a single image or uncalibrated multi-view images. Specifically, it proposes a self-evolved signed distance module (SeSDF), which refines the signed distance field derived from SMPL-X using both 2D pixel-aligned and space-aligned 3D image features. This approach enhances the signed distance field with additional clothing information that is consistent with the image features:
\begin{equation}
\begin{aligned}
    F_{\text{sdf}}\left(\mathcal{T}(\varphi(S), \mathbf{x}), \mathcal{B}(\psi(I), \pi(\mathbf{x})), \mathcal{D}\left(d(\mathbf{x})\right), \mathbf{n}(\mathbf{x})\right) \mapsto(d', \mathbf{n}'),
\end{aligned}
\end{equation}
\begin{equation}
    \mathcal{D}(d)=(d, \sin (2^0 \pi d), \cos (2^0 \pi d), \ldots, \sin (2^L \pi d), \cos (2^L \pi d)),
\end{equation}
where $d' \in \mathbb{R}$ denotes the SDF derived from the SMPL-X model.
Afterwards, given a 3D point $X$ and its features, the implicit function of SeSDF can be formulated as:
\begin{equation}
\begin{aligned}
    F\left(\mathcal{T}(\varphi(S), \mathbf{x}), \mathcal{B}(\psi(I), \pi(\mathbf{x})), \mathcal{D}(d'(\mathbf{x})), \mathbf{n}'(\mathbf{x}), z(\mathbf{x})\right) \mapsto s.
\end{aligned}
\end{equation}

\textbf{ECON}
Summarizing the above implicit methods reveals two challenges: using only image information can cause depth ambiguity and human pose inaccuracy, while incorporating 3D priors often results in missing clothing details as SMPL or SMPL-X contains only minimal clothes. 
Consequently, ECON~\cite{xiu2023econ} introduces an explicit method instead of the implicit function. It first follows ICON to predict 2D front and back normal and depth maps from the input image and the predicted SMPL-X model. It then uses a depth-aware silhouette-consistent bilateral normal integration (d-BiNI) optimizer~\cite{cao2022bilateral} to recover the 3D front and back surfaces separately. Based on these partial surface estimates, it applies IF-Nets+~\cite{chibane2020implicit} to implicitly complete the body. With optional face or hands from the SMPL-X model, screened poisson~\cite{kazhdan2013screened} is finally employed to combine all the 3D surfaces to form the complete 3D human body.

Other works including IP-Net~\cite{bhatnagar2020combining}, S-PIFu~\cite{chan2022s}, and IntegratedPIFu~\cite{chan2022integratedpifu} also incorporate priors like SMPL model and depth for better human reconstruction, which readers can refer to on their own for detailed illustration.

\subsubsection{Qualitative and Quantitative Evaluations}

\begin{table*}[h]
\scriptsize
\centering
\caption{Quantitative comparison on the single-view setting. Results obtained from~\cite{xiu2023econ}.}
\resizebox{1.\textwidth}{!}{
\begin{tabular}{l||ccccc||cccccc}
\toprule[1pt]
{\multirow{3}*{Method}} & \multicolumn{5}{c||}{\multirow{2}*{Feature Included}} & \multicolumn{6}{c}{Quantitative Number} \\
{} & \multicolumn{5}{c||}{} & \multicolumn{3}{c}{\emph{CAPE}~\cite{ma2020learning}}  & \multicolumn{3}{c}{\emph{RenderPeople}~\cite{RenderPeople}} \\
\cmidrule[0.5pt](rl){2-6}
\cmidrule[0.5pt](rl){7-9}
\cmidrule[0.5pt](rl){10-12}
{} & {2D Feature} & {3D Feature} & {SMPL(-X)} & {Normal} & {Method Type} & {Chamfer$\downarrow$}& {P2S$\downarrow$} & {Normals$\downarrow$} & {Chmafer$\downarrow$} & {P2S$\downarrow$} & {Normals$\downarrow$} \\
\midrule[1pt]
PIFu~\cite{saito2019pifu} & \checkmark & & & & implicit & 1.722 & 1.548 & 0.0674 & 1.706 & 1.642 & 0.0709 \\
PIFuHD~\cite{saito2020pifuhd} & \checkmark & & & & implicit & 3.767 & 3.591 & 0.0994 & 1.946 & 1.983 & 0.0658 \\
PaMIR~\cite{zheng2021pamir} & \checkmark & \checkmark & \checkmark & \checkmark & implicit & 0.989 & 0.992 & 0.0422 & 1.296 & 1.430 & 0.0518 \\
ICON~\cite{xiu2022icon} & \checkmark & & \checkmark & \checkmark & implicit & 0.971 & 0.909 & 0.0409 & 1.373 & 1.522 & 0.0566 \\
ECON~\cite{xiu2023econ} & & & \checkmark & \checkmark & explicit & 0.996 & 0.967 & 0.0413 & 1.401 & 1.422 & 0.0516 \\
\bottomrule[1pt]
\end{tabular}}
\label{tab:pifucomparison}
\end{table*} 

\begin{figure}[t]
  \centering
  \includegraphics[width=8.5cm]{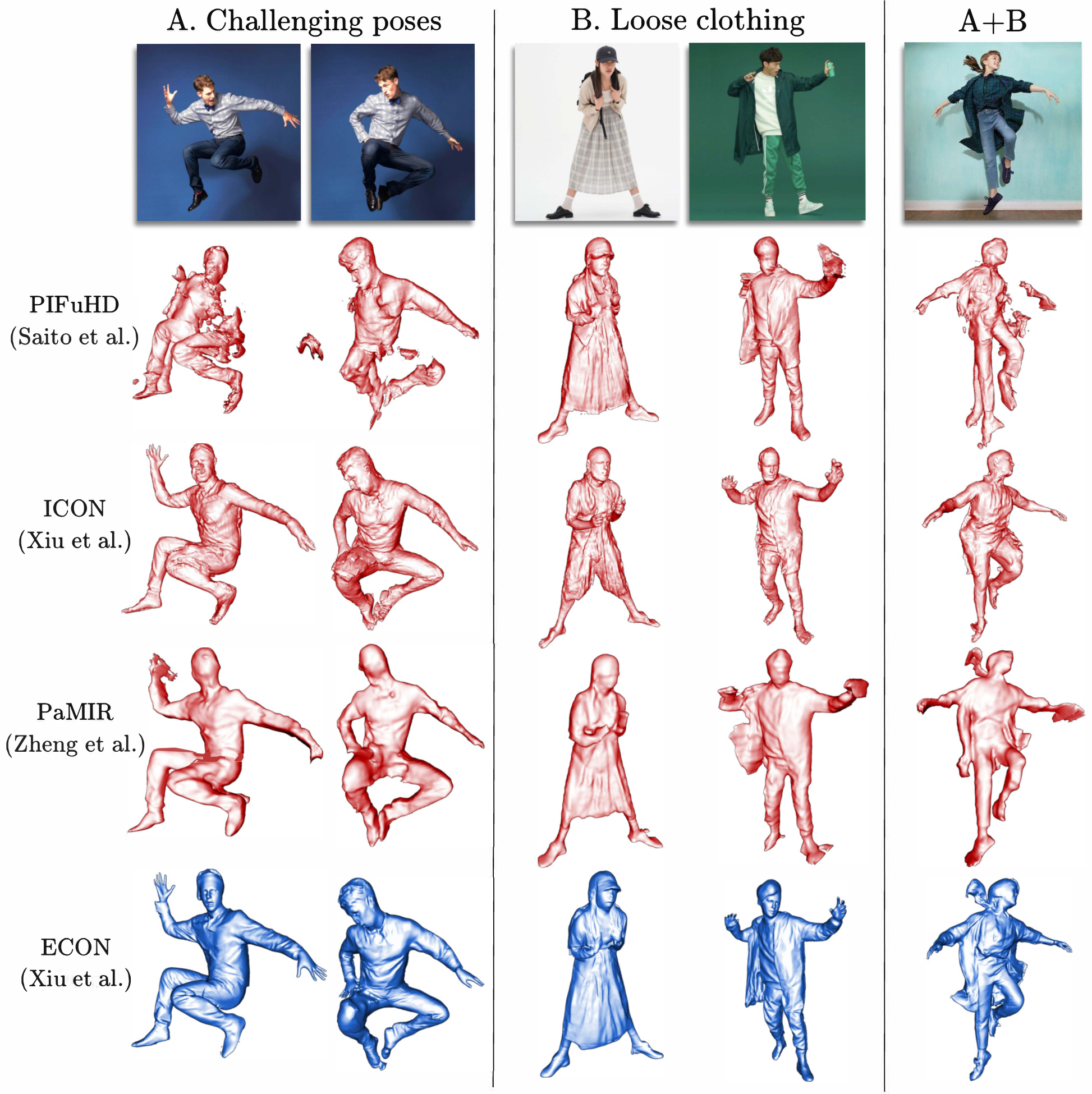}\\
  \caption{Qualitative comparison on the single-view setting. Figure obtained from \cite{xiu2023econ}.}
  \label{fig:pifucomparison}
\end{figure}

We first select typical implicit function-based methods that use only 2D features (PIFu~\cite{saito2019pifu}, PIFuHD~\cite{saito2020pifuhd}), as well as PaMIR~\cite{zheng2021pamir}, ICON~\cite{xiu2022icon} that utilize 3D features, and the explicit method ECON~\cite{xiu2023econ} for quantitative evaluations. By referring to~\tref{tab:pifucomparison}, we can observe that the incorporation of 3D priors leads to largely enhanced accuracy in capturing the human poses, resulting in much lower Chamfer, Point-to-Surface (P2S) and normal errors for both CAPE~\cite{ma2020learning} and RenderPeople~\cite{RenderPeople} test sets. We further provide qualitative evaluations in~\fref{fig:pifucomparison}. Due to the reliance on the SMPL/SMPL-X model, ICON and PaMIR present degenerated clothing details than PIFuHD which includes only 2D image features. 

Among them, ECON~\cite{xiu2023econ} introduces an explicit method that is capable of maintaining the robustness of explicit shape models for unseen poses without sacrificing the topological flexibility of implicit functions for loose clothing. However, it is important to note that single-view image only provides information about the front view of the human body, which inherently limits the effectiveness of these reconstruction methods.

\subsection{Multi-view Scenarios}
\label{sec: pifu-multiview}

Multi-view images can offer a more comprehensive set of 3D information that can compensate each other for getting better reconstruction results in terms of (1) finer clothing details, and (2) enhanced accuracy in capturing motion.
However, the key challenge lies in effectively integrating features from different viewpoints, especially when occlusions occur in multiple views or when information for the same 3D query point varies across viewpoints. 

PIFu~\cite{saito2019pifu} incorporates multi-view features through average pooling that treats multi-view features equally.
However, this approach overlooks the different quality of predictions from different views. For example, the image feature of a 3D point extracted from a non-occluded view likely yields the most accurate prediction, while features extracted from an occluded or lateral view should have minimal impact on the prediction. Obviously, average pooling is not an optimal method for feature fusion.

Researchers therefore explore different feature fusion strategies. StereoPIFu \cite{hong2021stereopifu} considers a pair of stereo images as input for depth-aware reconstruction, DeepMultiCap~\cite{zheng2021deepmulticap} leverages a self-attention mechanism for multi-view fusion, SeSDF~\cite{cao2023sesdf} proposes an occlusion-aware feature fusion strategy to fuse features from different views effectively, DiffuStereo \cite{shao2022diffustereo} introduces a multi-level diffusion-based stereo network to produce highly accurate depth maps, which are then converted into a high-quality 3D human model through an efficient multi-view fusion strategy.

\begin{figure}[t]
  \centering
  \includegraphics[width=8.5cm]{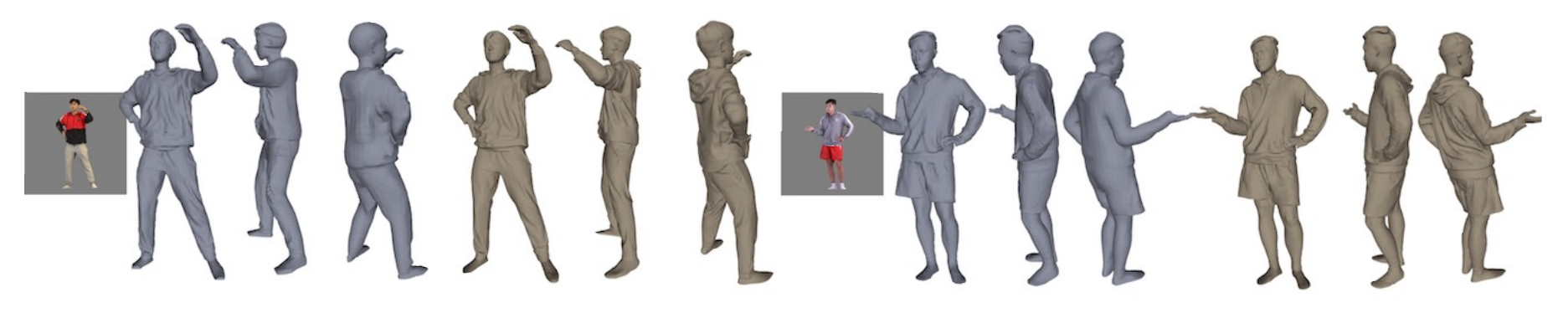}\\
  \caption{Single-view reconstruction (grey) vs multi-view reconstruction (yellow). Figure obtained from~\cite{cao2023sesdf}.}
  \label{fig:multi-view1}
\end{figure}

\begin{figure}[t]
  \centering
  \includegraphics[width=8.5cm]{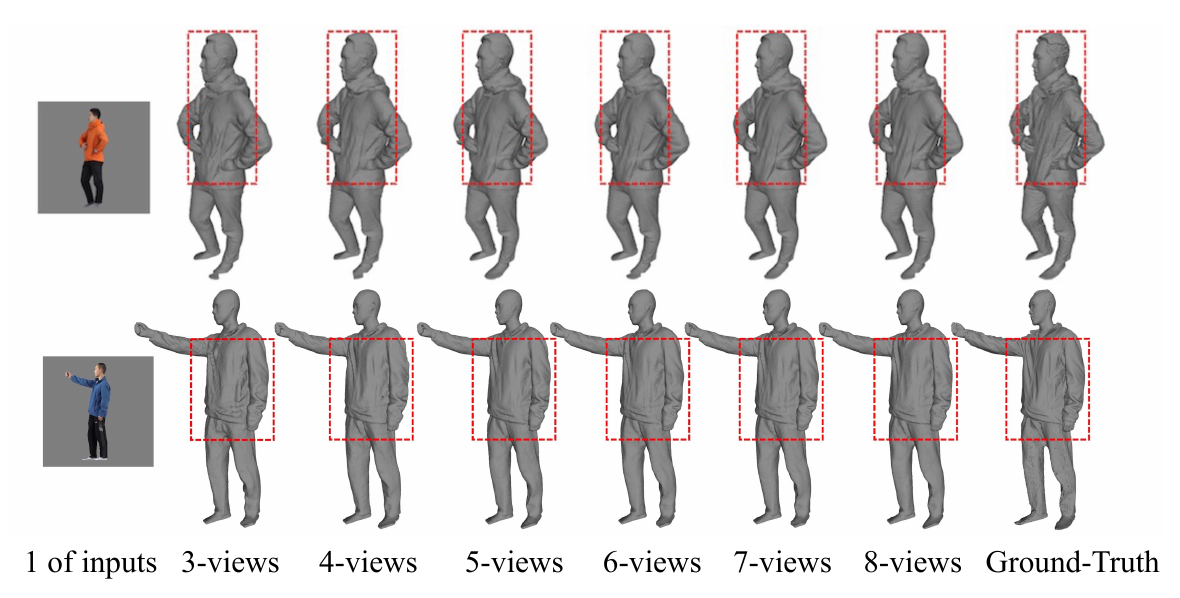}\\
  \caption{Reconstruction with different numbers of input views. Figure obtained from~\cite{cao2023sesdf} and model is trained with three views.}
  \label{fig:multi-view2}
\end{figure}

\subsubsection{Qualitative and Quantitative Evaluations}

We begin by presenting a comparison between single-view and multi-view 3D human reconstruction and show the visualizations in~\fref{fig:multi-view1}. Based on these qualitative results, we observe that multi-view 3D human reconstruction consistently achieves more accurate 3D human poses and exhibits enhanced clothing details by leveraging additional information in the 3D space.

To further evaluate the impact of the number of input images, we present qualitative assessments in~\fref{fig:multi-view2}. By examining these visualizations, we can observe an enhancement in the reconstruction quality when the number of input views increased from three to five. Nevertheless, the enhancement might become minimal while the number of views continues to increase. This situation primarily arises due to two factors: (1) the input views are already saturated, and (2) the performance might be correlated with the number of views used in model training.

% (2) The performance of the reconstruction is also influenced by the number of images used during the training phase. Optimal reconstruction results are likely to be achieved when testing with the same number of images as used in training.

%% file: Sections/3_NeRF-based/NeRF-based.tex
\section{NeRF-based 3D Human Novel View Synthesis}
\label{sec:nerf}

\subsection{Neural Radiance Fields (NeRF)}

\begin{figure}[t]
  \centering
  \includegraphics[width=8.5cm]{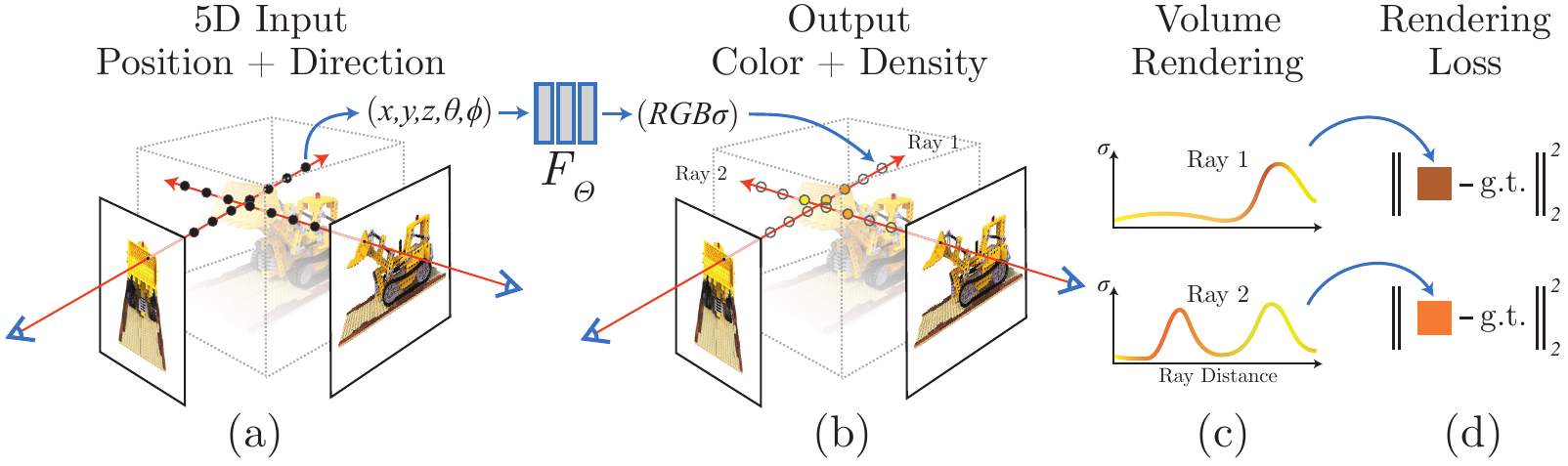}\\
  \caption{NeRF overview. Figure obtained from~\cite{mildenhall2021nerf}.}
  \label{nerf-overview}
\end{figure}

Unlike 3D implicit human reconstruction methods that show high demands on the quantity and quality of the 3D hard-to-obtain dataset, NeRF~\cite{mildenhall2021nerf} can achieve photo-realistic novel-view synthesis using only a limited set of images. Readers please refer to~\fref{nerf-overview} for the overall framework of NeRF.

NeRF represents a 3D scene via an implicit function:
\begin{equation}\label{nerf}
    F_{\Theta}(\gamma(\mathbf{x})) \mapsto (\mathbf{c}, \sigma),
\end{equation}
where $\gamma(\cdot)$ is a grid frequency encoder that lifts a 3D point $\mathbf{x}$ to a higher dimension, $\mathbf{c}$ is the RGB color, and $\sigma$ is the density value. Generally, the implicit function $F_{\Theta}(\cdot)$ is implemented as an MLP with trainable parameters $\Theta$. We omit $\Theta$ in the remaining sections for simplicity.

After getting the density and RGB color of a 3D point, NeRF employs a differentiable volume rendering module to render a 3D scene onto a 2D image. For each image pixel from a certain camera angle, the rendering involves casting a ray $\mathbf{r}$ from the pixel location into the 3D scene and sampling 3D points $\boldsymbol{\mu}_i$ along the ray. The color $C$ of each image pixel is aggregated from the sampled points' color values $\mathbf{c}$:
\begin{equation}
    C(\mathbf{r})=\sum_i W_i \mathbf{c}_i, \quad W_i=\alpha_i \prod_{j<i}\left(1-\alpha_j\right),
\end{equation}
where $\alpha_i=1-\mathrm{e}^{-\sigma_i\left\|\boldsymbol{\mu}_i-\boldsymbol{\mu}_{i+1}\right\|}$.

The loss for training NeRF is the total squared error between the rendered and true pixel colors:
\begin{equation}
    \mathcal{L}=\sum_{\mathbf{r} \in \hat{R}}\left\|C_{\text{gt}}(\mathbf{r})-C(\mathbf{r})\right\|_2^2,
\end{equation}
where $\hat{R}$ is the set of rays in each batch.

\subsection{NeRF-based Methods}

Adopting the neural radiance field, many methods have been proposed for 3D human reconstruction and animation from unstructured photo sets under various scenarios.

\subsubsection{3D Human Reconstruction from Static Cameras}

We first examine the scenario of 3D human reconstruction from static cameras, where all cameras are fixed in specific positions to capture images or videos of the human body.

Unlike PIFu which processes human subjects with only one pose during both training and inference, NeRF-based human reconstruction aims to optimize a 3D scene based on multiple images with various poses and viewpoints. To bridge the connections across different poses and angles, deformation fields based on SMPL(-X)~\cite{SMPL:2015, SMPL-X:2019} are usually applied to map a point $\mathbf{x}_o$ from an observation space to a corresponding point $\mathbf{x}_c$ in the canonical space. This process involves two key parts: (1) articulated deformation that applies the inverse transformation of SMPL linear blend skinning function $\mathtt{LBS}(\cdot)$ as in \Eref{smpl} and \Eref{smplx}, and (2) non-rigid motion implemented as an MLP to learn the corrective offset:
\begin{equation}\label{lbs-nerf}
    \mathbf{x}_c=\mathbf{x}_c^{l b s}+\mathtt{MLP}_{\theta_{\mathrm{NR}}}\left(\gamma\left(\mathbf{x}_c^{l b s}\right)\right), \mathbf{x}_c^{l b s}=\mathcal{G}^{-1} \cdot \mathbf{x}_o,
\end{equation}
where $\gamma(\cdot)$ is the grid frequency encoder proposed in NeRF, and $\mathcal{G}$ annotated in \Eref{eq:deformation} is obtained from the observed SMPL vertex closet to $\mathbf{x}_o$.

\begin{figure}[t]
  \centering
  \includegraphics[width=8.5cm]{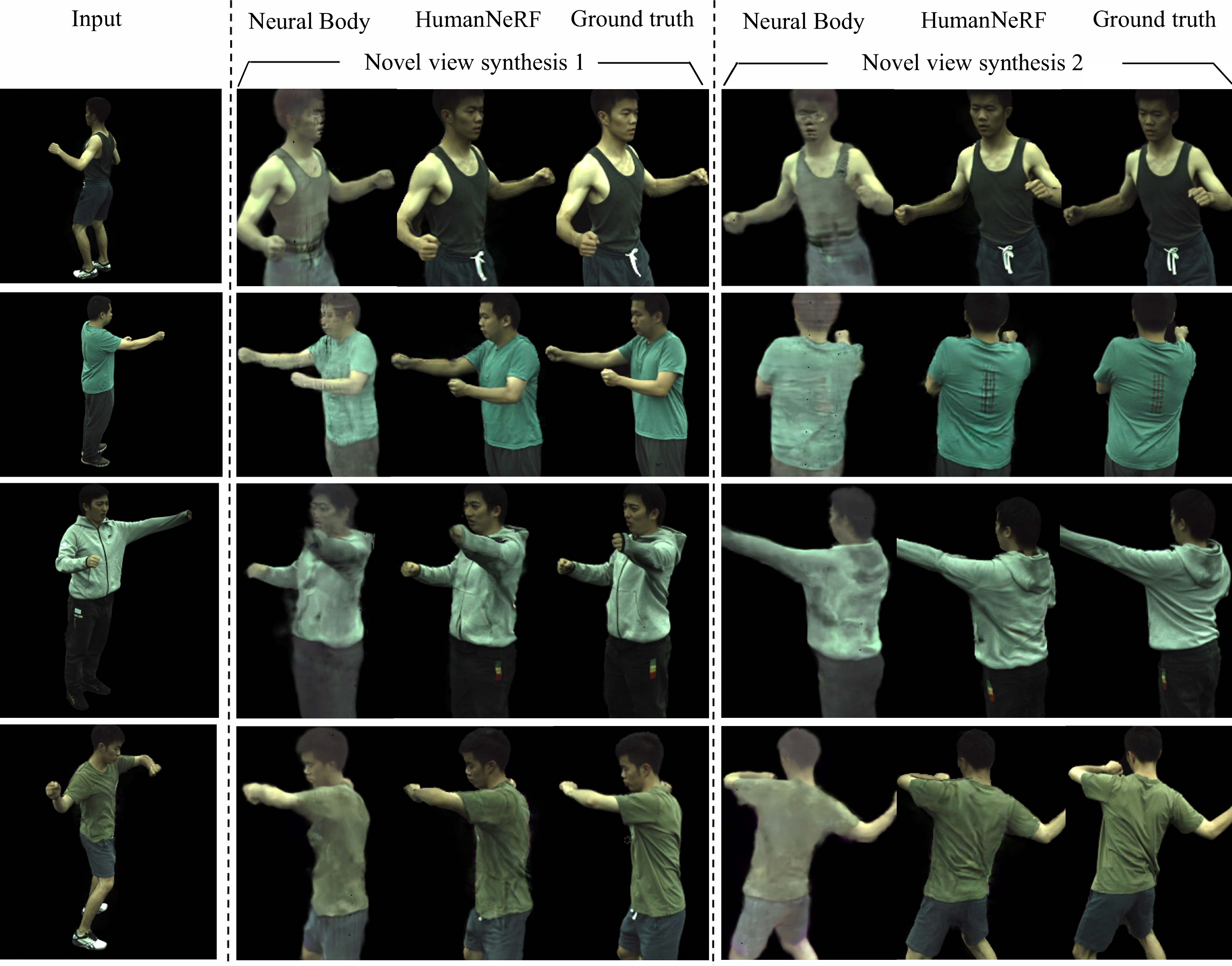}\\
  \caption{Input and output of NeRF-based reconstruction methods from static cameras. Figure obtained from~\cite{weng2022humannerf}.}
  \label{fg:nerf-reconstruction}
\end{figure}

To provide readers with a general understanding of the setups employed in NeRF-based reconstruction methods from static cameras, we present examples of their input and output in \fref{fg:nerf-reconstruction}.

\textbf{HumanNeRF}
HumanNeRF~\cite{weng2022humannerf} is one of the first to apply the deformation field to capture dynamic human models from monocular images under various viewpoints and motions. Specifically, it maps each point $\mathbf{x}_o$ sampled during volume rendering to its canonical counterpart $\mathbf{x}_c$ via \Eref{lbs-nerf} before calculating the density and color value:
\begin{equation}
    F(\gamma(\mathbf{x}_c)) \mapsto (\mathbf{c}, \sigma_c).
\end{equation}
Considering that the SMPL poses are separately estimated for each image (which may not be consistent), it further refines the body pose during training.

\textbf{Neural Body}
Neural-Body~\cite{peng2021neural} advances HumanNeRF by introducing structured latent codes anchored to the SMPL model to provide regularization. These latent codes, $\mathcal{Z}=\{z_1, z_2, \dots, z_{6890}\}$, correspond to SMPL's $6890$ vertices. To address their sparsity in 3D space, Neural-Body uses SparseConvNet~\cite{graham20183d} to process the latent codes and obtain 3D voxel-based feature volume. For any point $\mathbf{x}$, it is first transformed into the SMPL coordinate system, which aligns the point with the latent code volume. The latent code $\psi\left(\mathbf{x}, \mathcal{Z}, S_t\right)$ is then computed by tri-linear interpolation, and then input to two MLPs $F_{\sigma}$ and $F_{\mathbf{c}}$ to estimate the density and color:
\begin{subequations}
    \begin{gather}
        F_{\sigma}\left(\psi\left(\mathbf{x}, \mathcal{Z}, S_t\right)\right) \mapsto \sigma, \\
        F_{\mathbf{c}}\left(\psi\left(\mathbf{x}, \mathcal{Z}, S_t\right), \gamma_{\mathbf{d}}(\mathbf{d}), \gamma_{\mathbf{x}}(\mathbf{x}), \ell_t\right) \mapsto \mathbf{c},
    \end{gather}
\end{subequations}
where $S_t$ is SMPL parameters at frame $t$, $\gamma_{\mathbf{d}}$ and $\gamma_{\mathbf{x}}$ are positional encoding functions for viewing direction $\mathbf{d}$ and spatial location $\mathbf{x}$, and $l_t$ is the latent embedding.

\textbf{Neural Human Performer}
Unlike previous methods, Neural Human Performer~\cite{kwon2021neural} proposes a novel strategy for capturing information directly in the observation space. To achieve this, it first builds a skeletal feature bank by mapping the vertices of each SMPL model to its corresponding image and indexing the pixel-aligned image features. A temporal transformer then fuses these 2D features from different times (poses) and constructs a time-augmented skeletal feature bank. During the training and inference stages, Neural Human Performer (1) indexes skeletal features for a specific 3D point $\mathbf{x}$ from the feature bank using tri-linear interpolation, and (2) projects $\mathbf{x}$ to each image and acquires pixel-aligned image features via bi-linear interpolation. These features are then fused using a multi-view transformer to predict the density and color.

Besides the above methods: considering that multiple points in 3D space can be projected onto the same surface point on the mesh, Xu~\emph{et al.}~\cite{xu2022surface} utilizes barycentric interpolation with vertex normals to project points onto the mesh surface; I-M-Avatar~\cite{zheng2022avatar} presents a morphing-based implicit model tailored for head avatars and utilizes expression and pose deformations for detailed geometry and appearance; Semantic-Facial-NeRF~\cite{gao2022reconstructing} combines multi-level voxel fields with expression coefficients in the latent space to represent head avatars with complex facial attributes; Vid2Avatar~\cite{guo2023vid2avatar} proposes to jointly model the human and scene background via two neural radiance fields for a clean separation of the dynamic human and static background.

\subsubsection{3D Human Reconstruction from Dynamic Cameras}

\begin{figure}[t]
  \centering
  \includegraphics[width=8.5cm]{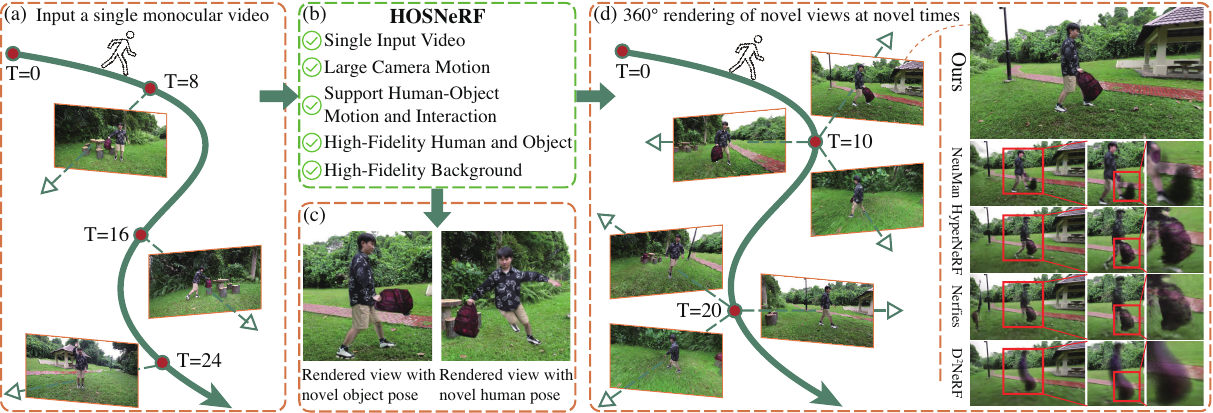}\\
  \caption{Input and output of HOSNeRF. Figure obtained from~\cite{liu2023hosnerf}.}
  \label{fg:hosnerf}
\end{figure}

Besides static camera scenarios, researchers also propose to reconstruct humans under dynamic camera settings where the cameras are in motion during the capture process, which are more suitable for real-world applications. See~\fref{fg:hosnerf} for an example of the input and output under such settings with dynamic cameras.

\textbf{HOSNeRF}
Considering that the SMPL model represents only the rough human body, disregarding the clothing topology and accessories, HOSNeRF~\cite{liu2023hosnerf} proposes object bones and state-conditional representations with learnable state embeddings for synthesizing 3D humans with items like bags. Specifically, object bones extend the human skeleton to better model deformations from human-object interactions. After defining the object bones, HOSNeRF also applies backward LBS and a non-rigid deformation module to map a 3D point from observation space to canonical space. In frame $i$, the dynamic 3D scene can be represented as:
\begin{equation}
    F(\gamma(\mathbf{x}_c), \mathcal{O}_c^i) \mapsto (\mathbf{c}, \sigma),
\end{equation}
where $\mathcal{O}_c^i$ is the learnable state embedding representing object states in the canonical space at frame $i$. As humans interact with objects at different times, state embeddings serve as conditions for learning representations of human-object interactions and the scene. Finally, HOSNeRF applies Mip-NeRF 360~\cite{barron2022mipnerf360} to represent the background scene.

Similar to HOSNeRF, NeuMan~\cite{jiang2022neuman} trains separate NeRF models for humans and scenes to reconstruct both from the input video. To further handle the challenge of maintaining identities through occlusion events, 4DHumans~\cite{goel2023humans} proposes an HMR 2.0 network, preserving more details in multiple-person scenarios. PPR~\cite{yang2023ppr} combines differentiable physics simulation and differentiable rendering via coordinate descent, largely reducing reconstruction artifacts. RAC~\cite{yang2023reconstructing} introduces a reconstruction technique for both animals and humans by learning skeletons with constant bone lengths within a video.

\subsubsection{3D Human Animation}\label{NARF}

\begin{figure}[t]
  \centering
  \includegraphics[width=8.5cm]{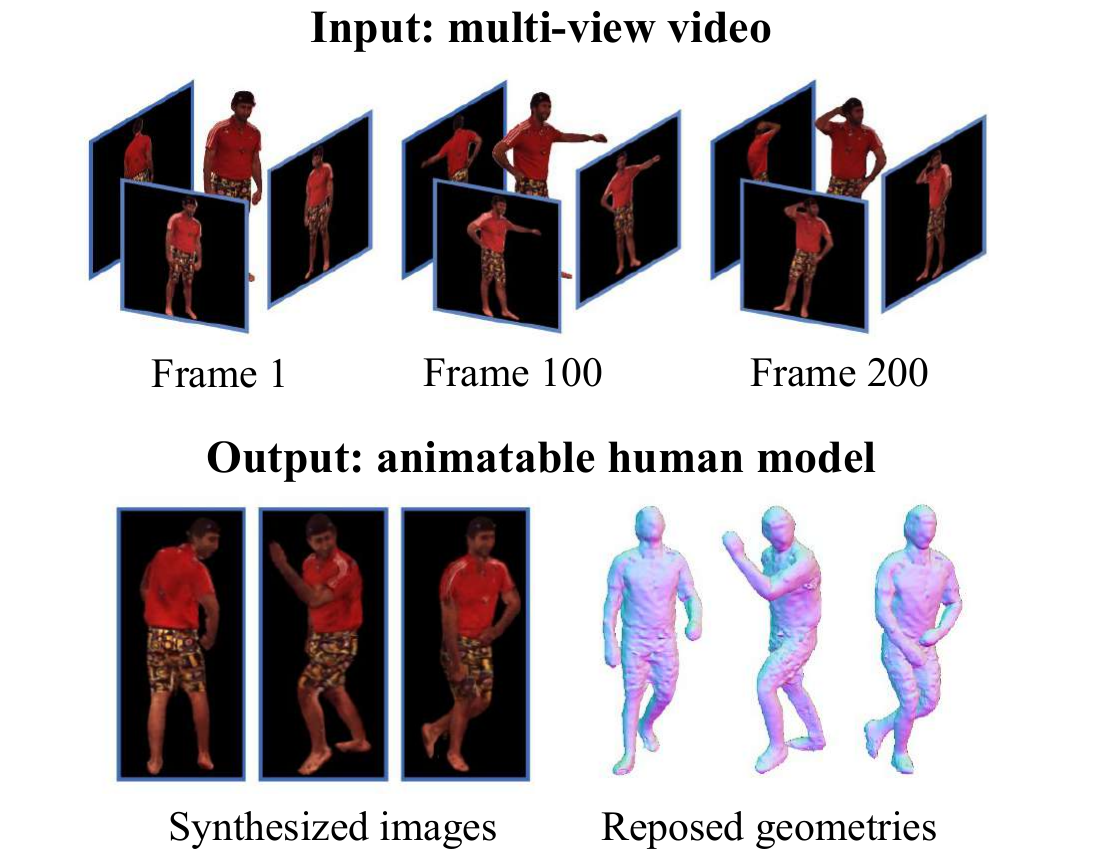}\\
  \caption{Input and output of NeRF-based animation methods. Figure obtained from~\cite{peng2021animatable}.}
  \label{fg:animatable-nerf}
\end{figure}

3D human animation focuses on learning how the clothing topology changes with different poses based on multi-view and multi-pose images. This enables the synthesis of free-viewpoint animations under user-guided pose sequences that are out of training distribution from multi-view videos. An example of the input and output of these animation methods is illustrated in~\fref{fg:animatable-nerf}.

\textbf{Neural Actor}
To achieve this, Neural Actor~\cite{liu2021neural} first adopts a deformation field that contains the articulate deformation and non-rigid motion (called residual deformation in Neural Actor), enabling the transfer of a 3D point $\mathbf{x}_o$ from the observation space to the canonical space $\mathbf{x}_c$. Additionally, it employs a feature extractor to acquire texture features from a 2D texture map, then indices the texture features of $\mathbf{x}_o$ based on its closest surface point $\mathbf{x}_s$:
\begin{equation}
    F(\gamma_{\mathbf{x}_c}(\mathbf{x}_c), \gamma_{\mathbf{d}}(\mathbf{d}), \mathcal{Z}(\mathbf{x}_s)) \mapsto (\mathbf{c}, \sigma),
\end{equation}
where $\mathcal{Z}(\cdot)$ denotes the extracted texture features. Specifically, during training, it obtains $\mathcal{Z}(\cdot)$ from the input multi-view images, while at the inference stage, it relies on a SMPL-based normal map to obtain $\mathcal{Z}(\cdot)$.

\textbf{Animatable NeRF}
Considering that articulated deformation may not efficiently capture clothing topology, Peng~\emph{et al.}~\cite{peng2021animatable} introduce a novel neural blend weight field to obtain better skinning weights $w_k$. To this end, they first define a per-frame latent code $\boldsymbol{l}_i$ that encodes the human appearance in frame $i$. Given a 3D point $\mathbf{x}$, they transform it to the canonical space by:
\begin{subequations}
    \begin{align}
        w_i^{'}(\mathbf{x}) &= \text{norm}(F_{\delta w}(\mathbf{x}, \boldsymbol{l}_i) + w_i(\mathbf{x}),\\
        \mathbf{x}_c &= (\sum_{k=1}^{K} w_k^{'}(\mathbf{x})\mathcal{G}_k)^{-1} \cdot \mathbf{x},
    \end{align}
\end{subequations}
where $F_{\delta w}$ is the neural blend weight field and the definition of other terms are the same as~\Eref{eq:deformation}. For each frame $i$, the color and density value are then given by:
\begin{subequations}
    \begin{gather}
        F_{\sigma}(\gamma_{\mathbf{x}}(\mathbf{x}_c) \mapsto (\sigma_i(\mathbf{x}_c), \mathbf{z}_i(\mathbf{x}_c)),\\
        F_{\mathbf{c}}(\mathbf{z}_i(\mathbf{x}_c), \gamma_{\mathbf{d}}(\mathbf{d}), \boldsymbol{l}_i) \mapsto \mathbf{c}_i(\mathbf{x}),
    \end{gather}
\end{subequations}
where $\mathbf{z}_i(\cdot)$ is the intermediate shape feature.

\textbf{TAVA}
Unlike previous methods that consider only one possible canonical point $\mathbf{x}_c$ after deformation, TAVA~\cite{li2022tava} proposes to find the canonical candidates $\{ \mathbf{x}_{c,1}, \mathbf{x}_{c,2}, ..., \mathbf{x}_{c,K} \}$ and predict the color and density values for all the candidates:
\begin{subequations}
    \begin{gather}
        F_{\mathbf{c}, \sigma}(\mathbf{x}_{c,i}, \boldsymbol{\Sigma}) \mapsto \mathbf{h} \mapsto (\mathbf{c}_{i}^{*}, \sigma_i),\\
        F_{a}(\mathbf{h}, \mathbf{d}) \mapsto a_i,
    \end{gather}
\end{subequations}
where $\boldsymbol{\Sigma}$ is the multivariate Gaussians applied in Mip-NeRF~\cite{barron2022mipnerf360}, $\mathbf{h}$ is the intermediate output, and $a_i$ is the ambient occlusion at point $\mathbf{x}_c$ under viewing direction $\mathbf{d}$. TAVA then chooses final values based on their density:
\begin{equation}
    \mathbf{c} = \mathbf{c}_{t}^{*} \cdot a_{t}, \quad \sigma = \sigma_t, \quad \text{where} \quad t = \argmax_i({\sigma_i}).
\end{equation}

\textbf{NARF}
Evolving from TAVA, NARF~\cite{2021narf} divides the articulated human into several rigid parts according to the skeleton of the SMPL model, and each rigid part represents a local coordinate system. For a 3D point $\mathbf{x}$, NARF first transforms it and the viewing direction $\mathbf{d}$ to each local system:
\begin{equation}
    \mathbf{x}^{i} = \mathbf{R}^{i^{-1}}(\mathbf{x} - \mathbf{t}^i), \quad \mathbf{d}^{i} = \mathbf{R}^{i^{-1}} \mathbf{d},
\end{equation}
where $\mathbf{R}^i$ and $\mathbf{t}^i$ are the corresponding rotation matrix and translation vector for part $i$. To predict the exact part that point $\mathbf{x}$ belongs to, NARF includes a selector network $\mathcal{S}$ that consists of $P$ lightweight sub-networks for each rigid part:
\begin{equation}
    F_{\mathcal{S}}(\gamma(\mathbf{x}^i, \gamma(\zeta)) \mapsto(s^i), \quad p^i = \frac{\text{exp}(s^i)}{\sum_{k=1}^{P}\text{exp}(s^k)},
\end{equation}
where $\zeta$ is the bone parameter, and $s^i$ is the occupancy value. The density and color can then be predicted via:
\begin{subequations}
    \begin{gather}
        F_{\sigma}(\{ \gamma_{\mathbf{x}}(\mathbf{x}^i) * p^i | i \in [1, P]\}, \gamma(\zeta)) \mapsto (\sigma, \mathbf{h}),\\
        F_{\mathbf{c}}(\mathbf{h}, \{(\gamma_{\mathbf{d}}(\mathbf{d}^i) * p^i, \gamma(\xi^i) * p^i | i \in [1, P] \} ) \mapsto \mathbf{c},
    \end{gather}
\end{subequations}
where $\xi^i$ is the 6D transformation vector of part $i$.

\textbf{ARAH}
Instead of applying the articulated deformation and non-rigid motion, ARAH~\cite{wang2022arah} introduces a joint root-finding algorithm designed to find a canonical point and its depth along the viewing direction, which satisfies both the SDF iso-surface condition\footnote{The canonical point should be on the iso-surface.} and the LBS condition\footnote{After applying forward LBS to the canonical point, the transformed point should lie on the specified camera ray.}. Subsequently, the SDF value and color networks employed in ARAH can be formulated as:
\begin{subequations}
    \begin{gather}
        F_{\text{SDF}}(\mathbf{x}_c, \theta, \beta, \boldsymbol{l}) \mapsto \mathbf{h} \mapsto d,\\
        F_{\mathbf{c}}(\mathbf{x}_c, \mathbf{n}_o, \mathbf{d}, \mathbf{h}, \boldsymbol{l}) \mapsto \mathbf{c},
    \end{gather}
\end{subequations}
where $\theta, \beta$ are the pose and shape parameters, $\boldsymbol{l}$ denotes the latent code as in Neural Body, $\mathbf{n}_o$ is the normal vector in the observation space, and $\mathbf{h}$ is the intermediate output for calculating the SDF value $d$.

\begin{figure}[t]
  \centering
  \includegraphics[width=8.5cm]{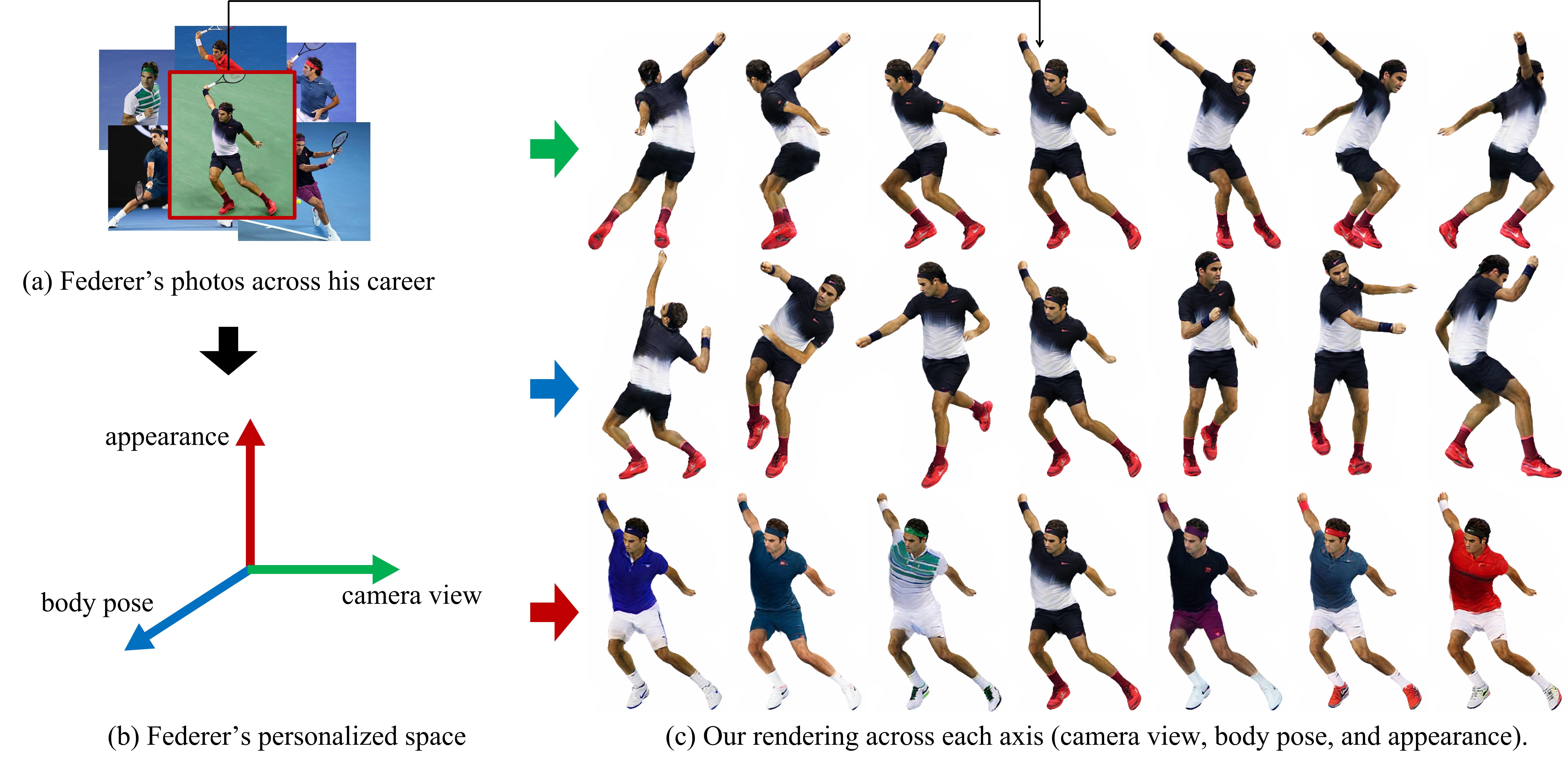}\\
  \caption{Input and output of PersonNeRF. Figure obtained from~\cite{weng2023personnerf}.}
  \label{fg:personnerf}
\end{figure}

\textbf{PersonNeRF}
Unlike previous methods that learn from videos and drive the human body following the given poses, PersonNeRF~\cite{weng2023personnerf} builds a neural radiance field spanned by camera view, body pose, and appearance based on an image collection of a specific person with different poses and clothes (see~\fref{fg:personnerf}). To achieve this and enable traversing the space to explore unobserved combinations of human attributes, PersonNeRF incorporates (1) a pose embedding to correlate the estimated pose, and (2) an appearance embedding pre-trained on the image collection to ensure appearance consistency. It further proposes to learn a shared motion field that outputs joint angle residuals and is learned across different clothings. It constrains all body poses in the dataset regardless of their appearance, enhancing pose generalization. 

Moreover,  several other works also leverage deformation fields for free-viewpoint animations.
MPS-NeRF~\cite{gao2022mps} introduces two deformation fields to connect the canonical space with the observation space and target space, one for extracting image features from the input image for radiance prediction and another for rendering the output image.
SHERF~\cite{hu2023sherf} utilizes 3D-aware global, point-level, and pixel-aligned features for effective encoding and a feature fusion transformer to predict the color and density.
MonoHuman~\cite{yu2023monohuman} proposes a Shared Bidirectional Deformation module to achieve generalizable consistent forward and backward deformation.
ActorsNeRF~\cite{mu2023actorsnerf} designs a 2-level canonical space (a category-level canonical space and an instance-level canonical space) for a coarse-to-fine strategy.

\subsubsection{Methods Combining Mesh and Radiance Field}

Following PIFu~\cite{saito2019pifu} and NeRF~\cite{mildenhall2021nerf}, many methods have also been proposed to combine the merits of both mesh and radiance fields for high-fidelity 3D human reconstruction.

\textbf{NeuralHumanFVV}
Among these methods, NeuralHumanFVV~\cite{suo2021neuralhumanfvv} consists of a neural geometry reconstruction stage and a neural blending stage to produce live 4D renderings based on 6-view dynamic videos. In the neural geometry reconstruction stage, it (1) extracts a coarse geometry prior via Shape-from-Silhouette (SfS)~\cite{cheung2003shape} algorithm, (2) obtains a finer geometry using a multi-view implicit function based on PIFu, and (3) utilizes a hierarchical sampling strategy to recover geometry details such as clothing folds. Specifically, in the hierarchical sampling strategy, a depth fine-tuning network takes the feature of the midpoint between two selected sample points $\mathbf{x}_1$ and $\mathbf{x}_2$ as input, and outputs the displacement of depth value, which is then used to refine the depth value. In the neural blending stage, it encodes the fine-detailed geometry and texture information from adjacent input views into a photo-realistic texture output.

\textbf{Function4D}
Function4D~\cite{tao2021function4d} achieves real-time reconstruction by integrating the proposed Dynamic Sliding Fusion (DSF) and deep implicit surface reconstruction. Unlike traditional volumetric fusion methods that attempt to complete surfaces by fusing all available temporal depth observations, DSF focuses on augmenting current observations to maintain consistency and reduce noise without relying on long-term tracking. This is done by confining tracking and fusion processes to a sliding window of the current, previous, and next frames. Having the depth map $\mathcal{D}$ extracted from DSF, the deep implicit surface reconstruction then includes (1) a GeoNet that uses truncated projective SDF (PSDF) values as a novel feature for preserving geometric details\footnote{$\mathbf{T}(\cdot)$ truncates the PSDF values in $[-\delta, \delta]$, with $\delta$ being a small positive threshold.}: 
\begin{equation}
    \text{PSDF}(\mathbf{x})=\mathbf{T}\left(d-\mathcal{B}(\pi(\mathbf{x}), \mathcal{D})\right),
\end{equation}
and (2) a ColorNet that utilizes a multi-head transformer network to aggregate features across multiple views.

\textbf{DoubleField}
Different from the above methods, DoubleField~\cite{shao2022doublefield} combines the surface and radiance field at the feature level in an implicit manner via a novel Network $F_{db}$. Given the query point $\mathbf{x}$, viewing direction $\mathbf{d}$ and 2D image features obtained from the input image $I$, $F_{db}$ learns a shared double embedding and predicts the occupancy $s$, the density $\sigma$ and the color $\mathbf{c}$ simultaneously. DoubleField network consists of a shared MLP (the Double MLP $F_{db}$) for learning the double embedding $e_{db}$ and two individual MLPs (the geometry MLP $F_g$ and the texture MLP $F_c$) for the surface and radiance fields prediction:
\begin{subequations}
\begin{gather}
F_{db}(\gamma(\mathbf{x}), \mathcal{B}(f_{2D}, \pi(\mathbf{x}))) \mapsto e_{db}, \\
F_g\left(e_{db}\right) \mapsto (s, \sigma), \quad F_c\left(e_{db}, d\right) \mapsto \mathbf{c}.
\end{gather}
\end{subequations}
Given that $s$ and $\sigma$ are both outputs from the same MLP layer, this approach inherently creates a robust link between the two fields, enabling their cooperation at the feature level.

\begin{figure}[t]
  \centering
  \includegraphics[width=8.5cm]{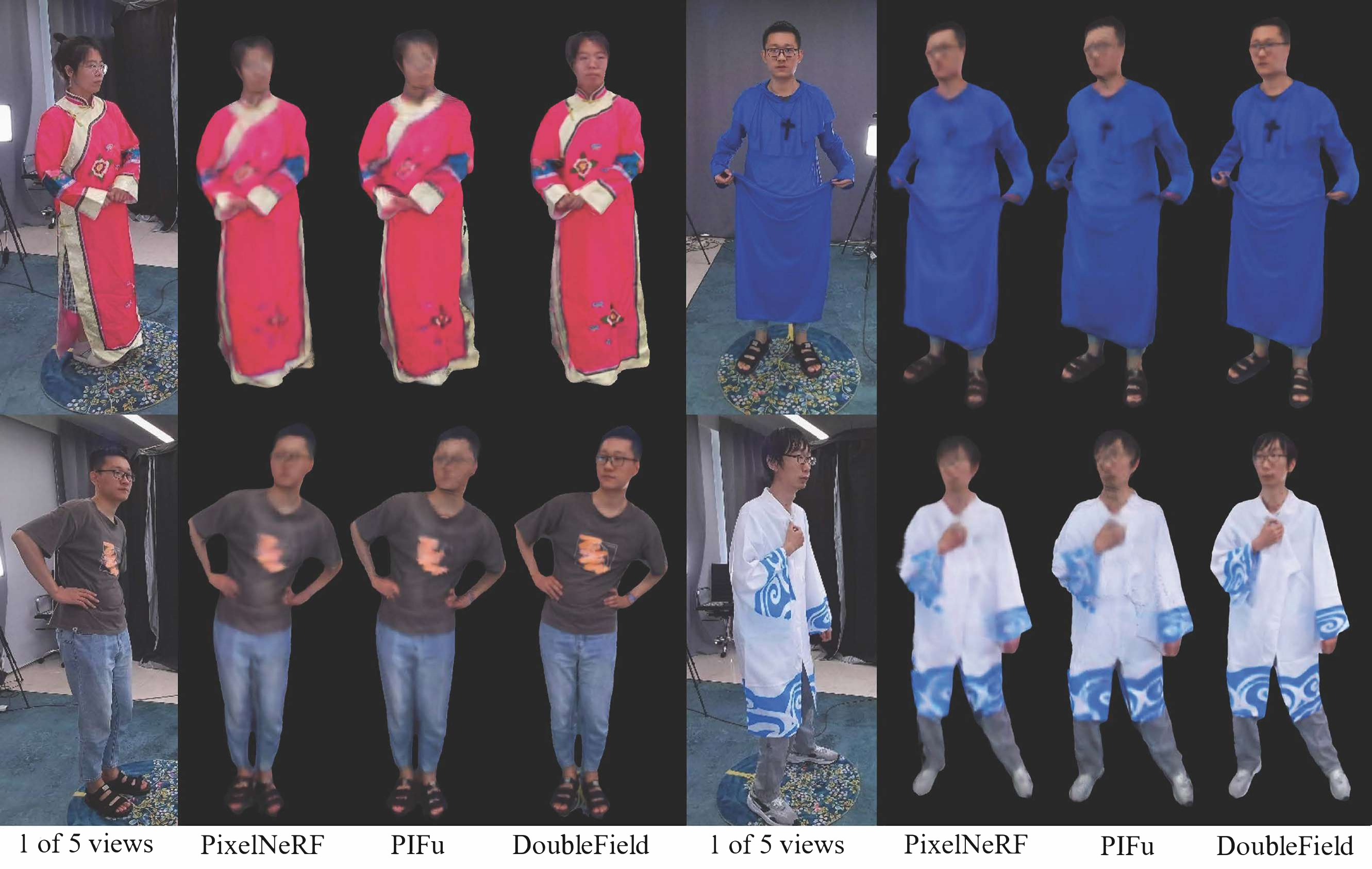}\\
  \caption{Comparison of methods combining mesh and radiance field with those using either alone (PIFu~\cite{saito2019pifu}: mesh only; PixelNeRF~\cite{yu2021pixelnerf}: radiance field only). Figure obtained from~\cite{shao2022doublefield}.}
  \label{fg:doublefield}
\end{figure}

Beyond the above-discussed methods, SelfRecon~\cite{jiang2022selfrecon} combines explicit and implicit geometry representations to obtain coherent geometry. DNA-Net~\cite{vo2023dna} includes a Neural Articulations Prediction Network (NAP-Net) to project observed points into the canonical space for improved learning of the geometry (SDF) and color fields.

In order to showcase the advancements achieved by methods that integrate both mesh and neural radiance field, we compare DoubleField with PIFu and PixelNeRF~\cite{yu2021pixelnerf} in~\fref{fg:doublefield}. This comparison highlights the superior performance of DoubleFiled, as it effectively combines the strengths of both representations to enhance the modeling of geometry and texture.

%% file: Sections/4_Gaussian/Gaussian.tex
\section{3D Gaussian-based Methods}
\label{sec:gaussian}

\subsection{3D Gaussian Splatting}

In contrast to NeRF~\cite{mildenhall2021nerf}, which employs neural networks to synthesize novel views, 3D Gaussian Splatting (3DGS)~\cite{kerbl20233d} introduces a novel approach that directly optimizes the position and attributes of 3D Gaussians, i.e, 3D position, opacity $\alpha$, anisotropic covariance, and spherical harmonic (SH)~\cite{ramamoorthi2001efficient} coefficients\footnote{Spherical Harmonic (SH) is used for controlling the color of each Gaussian to accurately capture the view-dependent appearance of the scene.}. This methodology enables us to efficiently represent and render intricate scenes at high resolutions, significantly reducing training time.

\emph{(1) Definition.}
3D Gaussian Splatting takes a set of static scene images and corresponding camera parameters obtained from Structure-from-Motion (SfM)~\cite{snavely2006photo} as inputs. For each SfM point $\mathbf{x}$, a 3D Gaussian is defined by a 3D covariance matrix $\boldsymbol{\Sigma}$ centered at point (mean) $\mu$:
\begin{equation}
    G(\mathbf{x})=e^{-\frac{1}{2}(\mathbf{x}-\mu)^T \boldsymbol{\Sigma}^{-1}(\mathbf{x}-\mu)}.
\end{equation}

\emph{(2) Rendering.}
To render the scene consisting of 3D Gaussians onto 2D image space, 3DGS incorporates the splatting rasterization.

a) Inspired by~\cite{lassner2021pulsar}, 3DGS designs a tile-based rasterizer. The screen is first divided into tiles (e.g., $16 \times 16$ pixels), and each Gaussian is instantiated according to the number of tiles they overlap and assigned a key that records view space depth and tile ID. Gaussians are then sorted based on their depth, allowing the rasterizer to correctly handle occlusions and overlapping geometry. 

b) 3DGS  introduces a point-based $\alpha$-blend rendering to obtain the RGB color $\mathbf{C}$. Specifically, it samples points along the ray with intervals $\delta_i$:
\begin{equation}
    \mathbf{C}_{\text {color }}=\sum_{i \in N} \mathbf{c}_i \alpha_i \prod_{j=1}^{i-1}\left(1-\alpha_j\right),
\end{equation}
with
\begin{equation}
    \alpha_i=\left(1-\exp \left(-\sigma_i \delta_i\right)\right),
\end{equation}
where $\sigma_i$ and $\mathbf{c}_i$ are the density and color of each point along the ray.

\emph{(3) Optimization.}
Following~\cite{fridovich2022plenoxels, sun2022direct}, Stochastic Gradient Descent techniques are utilized for optimization.
Specifically, 3DGS calculates the initial covariance matrix as an isotropic Gaussian with axes equal to the average distance to the closest three points.
They use a standard exponential decay scheduling method similar to Plenoxels~\cite{fridovich2022plenoxels} but for positions only. The optimization procedure is supervised by calculating the $\mathcal{L}_1$ loss and the D-SSIM term between the ground truth $g$ and the rendering $r$:
\begin{equation}
    \mathcal{L}=(1-\lambda) \mathcal{L}_1+\lambda \mathcal{L}_{\mathrm{D}-\mathrm{SSIM}},
\end{equation}
where
\begin{subequations}
\begin{gather}
    \mathcal{L}_1=\frac{1}{N} \sum_{i=1}^N\left|g_i-r_i\right|, \\
    \mathcal{L}_{\mathrm{D}-\mathrm{SSIM}}=1-\frac{1}{M} \sum_{j=1}^M \frac{\left(2 \mu_g \mu_r+c_1\right)\left(2 \sigma_{g r}+c_2\right)}{\left(\mu_g^2+\mu_r^2+c_1\right)\left(\sigma_g^2+\sigma_r^2+c_2\right)},
\end{gather}
\end{subequations}
where $\mu_g$ and $\mu_r$ are the average intensities of the ground truth and the rendered image within each of $M$ local windows\footnote{Local windows refer to segmented areas of an image which are used to analyze statistical properties locally.}, $\sigma_g^2$ and $\sigma_r^2$ are the variances of the ground truth and the rendered image, respectively, and $\sigma_{gr}$ is the covariance between them. Constants $c_1$ and $c_2$ help stabilize the division with small denominators.

\begin{figure}[t]
  \centering
  \includegraphics[width=8.5cm]{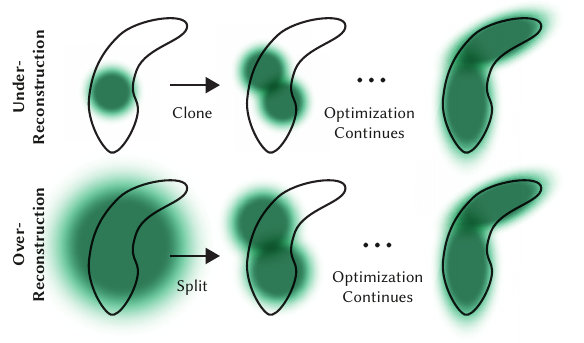}\\
  \caption{The densification and culling scheme. Figure obtained from~\cite{kerbl20233d}.}\label{densification}
\end{figure}

\emph{(4) Densification and Culling.}
To ensure that enough details and accurate reconstruction of the scene can be optimized, 3DGS incorporates densification and culling during the optimization. Specifically, 3DGS densifies every 100 iterations and remove any Gaussians that are essentially transparent, i.e., with opacity $\alpha$ less than a threshold $\epsilon_{\alpha}$. Readers can see~\fref{densification} for a detailed illustration of the densification scheme.

\begin{figure}[t]
  \centering
  \includegraphics[width=8.5cm]{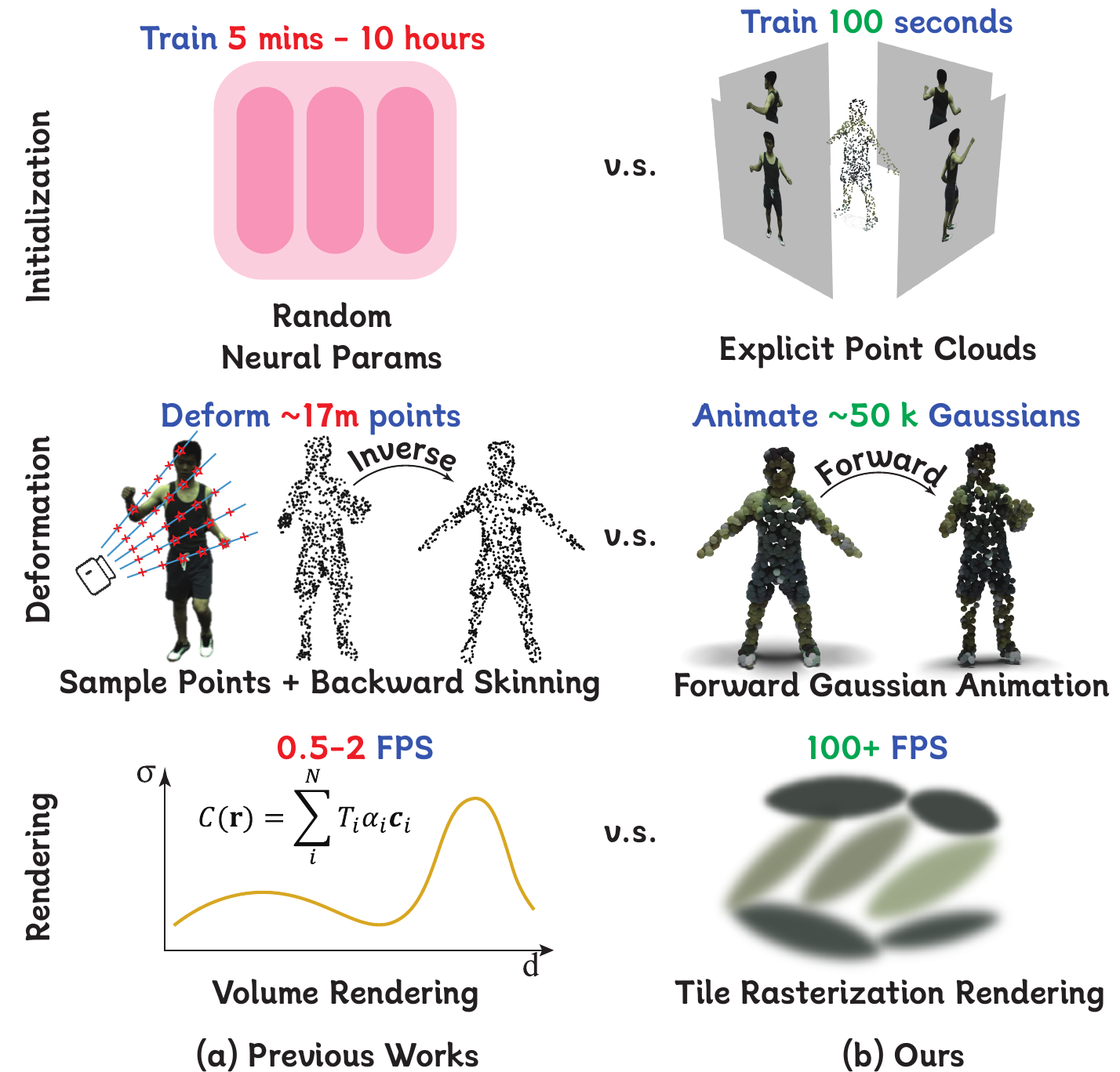}\\
  \caption{Differences between Gaussian-based methods and NeRF-based methods. Figure obtained from~\cite{li2023human101}.}\label{fg:difference}
\end{figure}

In order to provide readers with a clearer understanding of the differences between NeRF-based and Gaussian-based approaches, we have included a series of visualizations in Figure \ref{fg:difference}. By examining these visualizations, several key observations can be made: (1) NeRF-based methods employ MLPs to predict the density and color values while Gaussian-based methods directly optimize the attributes of 3D Gaussians. This direct optimization leads to significantly enhanced training efficiency. (2) Due to the disparity in their 3D representations, NeRF-based methods encounter non-correlated points, resulting in a larger number of points and significantly increased computational requirements. (3) Gaussian-based methods have the advantage of rendering the 3D scene directly through tile-based rasterization. This enables them to achieve greatly improved real-time performance.

\subsection{3D Human Reconstruction}\label{3D Gaussian reconstruction}

Leveraging 3DGS~\cite{kerbl20233d}, several works have been introduced for human-related tasks.

\textbf{Animatable 3D Gaussian}
Animatable-3D-Gaussian~\cite{liu2023animatable} is one of the first methods that apply 3DGS to enhance training efficiency and reduce GPU requirements. It incorporates: (1) A deformation network that transforms the point $\mathbf{x}_c$, rotation matrix $\mathbf{R}_c$, and viewing direction $\mathbf{d}_c$ from the canonical space to the observation space:
\begin{equation}
    \begin{gathered}
       \mathbf{x}_o = \sum_{i=1}^{n_b} w_i B_i^t \mathbf{x}_c, \\
       \mathbf{R}_o=\sum_{i=1}^{n_b} w_i B_i^t \mathbf{R}_c, \quad \mathbf{d}_o=\left(\sum_{i=1}^{n_b} w_i B_i^t\right)^{-1} \mathbf{d}_c, 
    \end{gathered} 
\end{equation}
where $n_b$ is the number of bones, $w_i$ is the skinning weight as in \Eref{eq:deformation}, and $B_i^t$ represents the transformation of the $i$-th bone in frame $t$, and (2) a time-dependent ambient occlusion that addresses the issue of dynamic shadows. Specifically, it predicts the ambient occlusion factor $a_o \in [0, 1]$ based on position $\mathbf{x}_o$ and hash-encoded time $t$:
\begin{equation}
    a_o = \mathtt{MLP}(\mathbf{x}_o, \gamma(t)), \quad \mathbf{c} = a_o * \mathbf{c}_o.
\end{equation}

\textbf{Drivable 3D Gaussian Avatars}
Drivable-3D-Gaussian-Avatars (D3GA)~\cite{zielonka2023drivable} leverages tetrahedral cages and cage-based deformation fields to model the body and individual garments, learning both the 3D human-related scenes and human segmentation maps. Given the canonical cage $\mathbf{v}$ and pose $\theta$, D3GA first employs two separate MLPs to model the deformation field of tetrahedron $i$:
\begin{subequations}
\begin{align}
& \Psi_{\mathtt{MLP}}:\left\{\theta, \gamma_{\mathbf{v}}(\mathbf{v})\right\} \rightarrow \Delta \mathbf{v}, \\
& \Pi_{\mathtt{MLP}}:\left\{\theta, \mathbf{b}_i, \mathbf{r}_i, \mathbf{s}_i\right\} \rightarrow\left\{\Delta \mathbf{b}_i, \Delta \mathbf{r}_i, \Delta \mathbf{s}_i \right\}, 
\end{align}
\end{subequations}
where the cage node correction network $\Psi_{\mathtt{MLP}}$ takes position-encoded canonical vertices $\gamma_{\mathbf{v}}(\mathbf{v})$ to predict offsets $\Delta \mathbf{v}$ for the cage node positions. The Gaussian correction network $\Pi_{\mathtt{MLP }}$ uses the canonical Gaussian parameters (barycentric coordinates $\mathbf{b}_i \in \mathbb{R}^4$, rotation $\mathbf{r}_i \in \mathbb{R}^4$ and scale $\mathbf{s}_i \in \mathbb{R}^3$ ) to predict their corrections. A shading network $\Gamma_{\mathtt{MLP}}$ is subsequently applied to learn the color $\mathbf{c}$ and opacity $o_i$:
\begin{equation}
    \Gamma_{\mathtt{MLP}}:\left\{\theta, \gamma_{\mathbf{d}}\left(\mathbf{d}_k\right), \mathbf{h}_i, \mathbf{f}_j\right\} \rightarrow\left\{\mathbf{c}_i, o_i\right\},
\end{equation}
where $\mathbf{h}_i$ is the auto-decoded~\cite{park2019deepsdf} feature vector of the initial color, $\mathbf{f}_j$ is the embedding vector with the time frame of the current sample.

\textbf{Human101}
Human101~\cite{li2023human101} reconstructs and animates dynamic human avatars from single-view videos with real-time rendering. Given an input video, it first extracts multiple sets of images with various poses. Each set contains four images of four orientations (front, back, left, right), which are used to create corresponding meshes using ECON~\cite{xiu2023econ}. The meshes are then deformed to the canonical space using inverse LBS from \Eref{smpl} and fused into a canonical point cloud. Following the original 3DGS, the point cloud is converted into canonical Gaussians for initialization. During animation, the Gaussians are deformed into the target pose by modifying their positions, rotations, and scales, and adjusting spherical harmonic coefficients. To address potential inconsistencies in human movement, Human101 employs an MLP to predict the residuals of position $\mathbf{x}$, rotation $\mathbf{r}$, and scale $\mathbf{s}$:
\begin{equation}
    F(\gamma_{\mathbf{x}}(\mathbf{x}), \gamma_t(t), \theta, \beta) \mapsto (\Delta \mathbf{x}, \Delta \mathbf{r}, \Delta \mathbf{s}).
\end{equation}

\textbf{Gaussian Head Avatar}
Gaussian-Head-Avatar~\cite{xu2023gaussian} proposes to reconstruct 3D heads based on 3DMM~\cite{blanz19993dmm}, 3D neutral landmarks, and triplane features. Specifically, it first construct a canonical neural Gaussian model with expression-independent attributes:
\begin{equation}
    \{ \mathbf{X}_o, \mathbf{F}_o, \mathbf{Q}_o, \mathbf{S}_o, \mathbf{A}_o \},
\end{equation}
where $\mathbf{X}_o \in \mathbb{R}^{N \times 3}$ is the position of the Gaussians in the canonical space, $\mathbf{F}_o \in \mathbb{R}^{N \times 128}$ denotes the point-wise feature vectors, $\mathbf{Q}_o \in \mathbb{R}^{N \times 4}, \mathbf{S}_o \in \mathbb{R}^{N \times 3}, \mathbf{A}_o \in \mathbb{R}^{N \times 1}$ represent the rotation, scale, and opacity respectively. To further improve the quality of geometry and texture, it proposes to condition the Gaussian attributes on the expression coefficients $\varepsilon_{\text{3DMM}}$ and head pose $\theta_{\text{3DMM}}$:
\begin{subequations}
    \begin{align}
        \mathbf{X} = &\mathbf{X}_o + \lambda_{\text{exp}}(\mathbf{X}_o) \cdot \mathtt{MLP}_{\mathbf{X}}^{\text{exp}}(\mathbf{X}_o, \varepsilon_{\text{3DMM}}) \notag\\ 
        &+ \lambda_{\text{pose}}(\mathbf{X}_o)\cdot \mathtt{MLP}_{\mathbf{X}}^{\text{pose}}(\mathbf{X}_o, \theta_{\text{3DMM}}), \\
        \mathbf{C} = &\lambda_{\text{exp}}(\mathbf{X}_o) \cdot \mathtt{MLP}_{\mathbf{C}}^{\text{exp}}(\mathbf{F}_o, \varepsilon_{\text{3DMM}}) \notag\\
        &+ \lambda_{\text{pose}}(\mathbf{X}_o)\cdot \mathtt{MLP}_{\mathbf{C}}^{\text{pose}}(\mathbf{F}_o, \theta_{\text{3DMM}}), \\
        \{\mathbf{Q}, \mathbf{S}, \mathbf{A} \} = &\{\mathbf{Q}_o, \mathbf{S}_o, \mathbf{A}_o \} + \lambda_{\text{exp}}(\mathbf{X}_o) \cdot \mathtt{MLP}_{att}^{\text{exp}}(\mathbf{F}_o, \varepsilon_{\text{3DMM}}) \notag\\
        &+ \lambda_{\text{pose}}(\mathbf{X}_o)\cdot \mathtt{MLP}_{att}^{\text{pose}}(\mathbf{F}_o, \theta_{\text{3DMM}}).
    \end{align}
\end{subequations}
By applying rigid rotations and translations to Gaussians, D3GA achieves dynamic 3D human head modeling in the observation space.

\textbf{SC-GS}
Given an image sequence from a monocular dynamic video, SC-GS~\cite{huang2023sc} utilize a set of sparse control points $\mathcal{P} = \{(p_i \in \mathbb{R}^3, \iota_i \in \mathbb{R}^{+})\}, \quad i\in\{1, 2, ..., N_p\}$\footnote{$p_i$ denotes the learnable coordinate of control point in the canonical space, and $\iota_i$ is the learnable radius parameter of a radial-basis-function (RBF) kernel that controls the impact of a control point on a Gaussian.} to reconstruct and drive 3D Gaussians for high-fidelity rendering. For each control point $p_i$, SC-GS first learns time-varying $6$ DoF transformations $[\mathbf{R}_i^t|\mathbf{T}_i^t] \in \mathbf{SE}(3)$, which consists of a local frame rotation matrix $\mathbf{R}_i^t \in \mathbf{SO}(3)$ and a translation vector $\mathbf{T}_i^t \in \mathbb{R}^3$:
\begin{equation}
    F(p_i, t) \mapsto (\mathbf{R}_i^t, \mathbf{T}_i^t).
\end{equation}
Given the learned 6 DoF transformations for the control points and for each 3D Gaussian, SC-GS applies k-nearest neighbor (KNN)~\cite{cover1967nearest} to obtain its $K(K=4)$ neighboring control points in the canonical space and calculate the interpolation weight:
\begin{equation}
    w_{jk} = \frac{\hat{w}_{jk}}{\sum_{k\in K} \hat{w}_{jk}}, \quad \text{where} \quad \hat{w}_{jk} = \text{exp}(-\frac{d_{jk}^2}{2o_k^2}),
\end{equation}
where $d_{jk}$ is the distance between the center of Gaussian $G_j$ and the neighboring control point $p_k$. By applying LBS~\cite{sumner2007embedded}, we can obtain the transformed Gaussian location $\mu^t_j$ and rotation matrix $R^t_j$ by:
\begin{subequations}
    \begin{align}
        \mu^t_j &= \sum_{k\in K}(\mathbf{R}^t_k(\mu_j - p_k) + p_k + \mathbf{T}^t_k), \\
        \mathbf{R}^t_j &= \left(\sum_{k \in K} w_{jk} r^t_k \right) \otimes \mathbf{R}_j,
    \end{align}
\end{subequations}
where $r^t_k$ is the quaternion of control point $k$ and $\otimes$ is the production process.

\textbf{GauHuman}
After initializing the position $\mathbf{x}_c$ of 3D Gaussians from SMPL vertex points, GauHuman~\cite{hu2023gauhuman} incorporates an LBS weight field module and a pose refinement module to transform Gaussians from the canonical space to the posed space. In the LBS weight field module, for each 3D Gaussian, GauHuman adds the LBS weight $w_k$ of nearest SMPL vertex with the predicted offsets $\mathtt{MLP}_{\Phi_{\mathrm{lbs}}}\left(\gamma\left(\mathbf{x}_c\right)\right)$ to predict the LBS weight coefficients $w'_k$:
\begin{equation}
    w'_k=\frac{\mathrm{e}^{\log \left(w_k+10^{-8}\right)+\mathtt{MLP}_{\text{lbs}}\left(\gamma\left(\mathbf{x}_c\right)\right)[k]}}{\sum_{k=1}^K \mathrm{e}^{\log \left(w_k+10^{-8}\right)+\mathtt{MLP}_{\text{lbs}}\left(\gamma\left(\mathbf{x}_c\right)\right)[k]}}.
\end{equation}
In the pose refinement module, it updates the joint angles $\theta^{\text{SMPL}}$ via another MLP:
\begin{equation}
    \theta=\theta^{\text{SMPL}} \otimes \mathtt{MLP}_{\text{pose}}(\theta^{\text{SMPL}}),
\end{equation}
where $\theta^{\text{SMPL}}$ is the SMPL body pose parameter estimated from images.
During optimization, GauHuman adopts the tile-based differentiable rasterizer from 3DGS for rapid rendering. They also introduce a strategy to adaptively control the number of Gaussians by employing human priors such as SMPL and Kullback-Leibler (KL) divergence to guide the split, clone, merge, and prune process.

Besides the methods mentioned above, Gaussian-Flow~\cite{lin2023gaussian} proposes a novel Dual-Domain Deformation Model for 4D scene training and rendering, avoiding per-frame 3DGS optimization. 3DGS-Avatar~\cite{qian20233dgs} develops a non-rigid deformation network to reconstruct human avatars, further enhancing the reconstruction speed. To capture finer details, GaussianBody~\cite{li2024gaussianbody} employs explicit pose-guided deformation to reduce ambiguity between the observation and the canonical space. GVA~\cite{liu2024gea} further introduces (1) a pose refinement method to enhance the alignment between the body and hand alignment, and (2) a surface-guided Gaussian re-initialization technique to address issues of unbalanced aggregation and initialization bias. There are also increasing efforts~\cite{guedon2023sugar, szymanowicz2023splatter, das2023neural, charatan2023pixelsplat} that utilize 3D Gaussian Splatting for object reconstruction. These papers can also provide valuable insights and inspiration for 3D human modeling.

%% file: Sections/5_GAN-based/GAN-based.tex
\section{GAN-based 3D Human Generation}
\label{sec:gan}

Moving to 3D human generation, we first discuss methods utilizing Generative Adversarial Network (GAN)~\cite{goodfellow2014generative}. Although StyleGAN~\cite{fu2022stylegan} and its follow-up works are conducted in the 2D image space, they provide the dataset and network basis for the recent 3D-GAN network. Therefore, this section mainly discusses StyleGAN-related techniques and 3D-GAN approaches that employ triplane~\cite{chan2022eg3d} as 3D representation.

\subsection{StyleGAN-related Techniques}

StyleGAN~\cite{fu2022stylegan} extends GAN~\cite{goodfellow2014generative} by introducing a "style" space to control human attributes. Its architecture contains two components: the mapping network and the generator.
Given the latent code $\mathbf{z}$, the mapping network is applied to predict a control signal $\mathbf{w}$. The generator takes $\mathbf{w}$ as the input to generate the final image via a sequential model comprised of several layers that progressively increase the resolution of the image. To enhance the influence of the control signal, $\mathbf{w}$ is used in each convolution layer of the generator after an adaptive instance normalization (AdaIN).

After StyleGAN, StyleGAN2~\cite{karras2020analyzing} and StyleGAN3~\cite{karras2021alias} are proposed to further enhance the image quality. StyleGAN2 introduces improvements in data augmentation and architectural changes. Specifically, it discards the progressive increase of layers for a more stable training process and replaces AdaIN with weight demodulation to reduce artifacts. StyleGAN3 is specifically designed to preserve image quality and distinctive features regardless of transformations. Even when an image is subject to rotation or zoom, the resulting modifications are still coherent and predictable.

StyleGAN also collects and annotates the Stylish-Humans-HQ Dataset (SHHQ), which contains over $230000$ high-quality images with various poses and textures. They introduce a "model zoo" that is trained on the SHHQ dataset using the StyleGAN2 framework and consists of six human-GAN models that are capable of generating full-body images with diverse poses and clothing textures. 

Leveraging the SHHQ dataset and these three models, many methods have been proposed to improve the performance from various perspectives. One such method is StyleSDF~\cite{or2022stylesdf}, which elevates the model to the 3D space by employing SDF-based volume rendering and a 3D implicit network. HoloGAN~\cite{nguyen2019hologan} learns  3D features from a 4D constant tensor and separates pose, shape, and appearance for finer details. BlockGAN~\cite{nguyen2020blockgan} learns 3D scene representations directly from unlabelled 2D images, providing control over each object’s 3D pose and identity. StylePeople~\cite{grigorev2021stylepeople} goes a step further by integrating the deformable SMPL-X model and neural rendering techniques into the StyleGAN2 network.

\subsection{3D-GAN Approaches}

\begin{figure}[t]
  \centering
  \includegraphics[width=8.5cm]{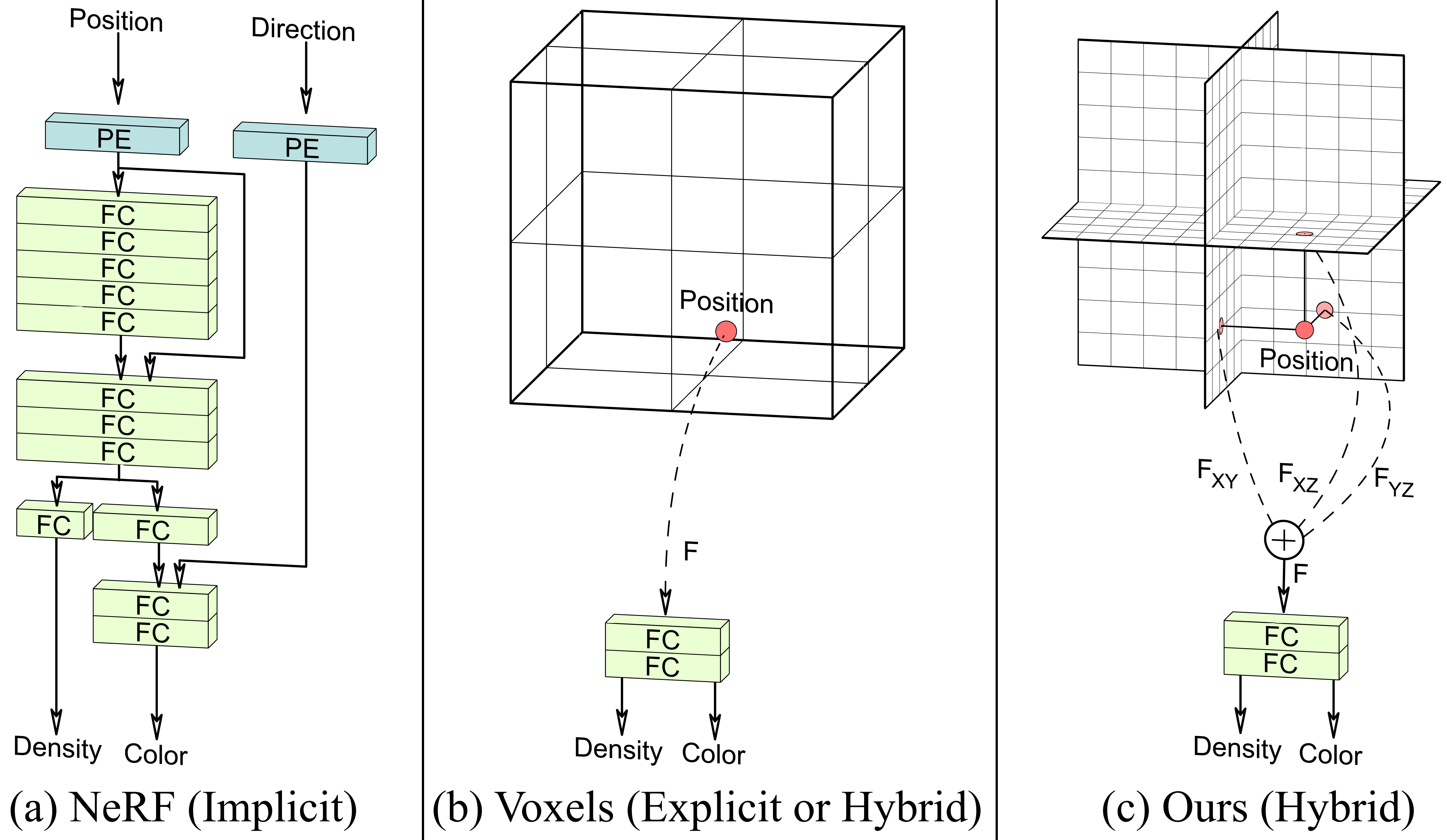}\\
  \caption{Triplane representation architecture. Figure obtained from \cite{chan2022eg3d}.}\label{triplane-overview}
\end{figure}

As NeRF~\cite{mildenhall2021nerf} has advanced the field, EG3D~\cite{chan2022eg3d} proposes a triplane representation and integrates it into the GAN framework. Remarkably, this approach represents one of the first instances where 3D model generation is accomplished using only 2D image data for training.  See ~\fref{triplane-overview} for the visualization of triplane and its difference with NeRF and voxel representation.

Specifically, the process begins by projecting each query point $\mathbf{x} \in \mathbb{R}^3$ onto three feature planes, enabling the indexing of the corresponding feature vectors $(f_{xy}, f_{xz}, f_{yz})$ through bilinear interpolation. These features are then summed, and input to an MLP to predict the color $\mathbf{c}$ and density $\sigma$:
\begin{equation}\label{eq:triplane}
    F(f_{xy}(\mathbf{x}) + f_{xz}(\mathbf{x}) + f_{yz}(\mathbf{x})) \mapsto (\sigma, \mathbf{c}).
\end{equation}
By incorporating volume rendering techniques, EG3D establishes a connection between the triplane representation and the StyleGAN2~\cite{karras2020analyzing} framework.

\subsection{3D Human Generation}

Following EG3D, many methods have been proposed to handle 3D human generation.

\textbf{GNARF}
To better present humans with dynamic motions, GNARF~\cite{bergman2022generative} proposes a Surface Field (SF) method to transform the query point $\mathbf{x}$ from the observation space to the canonical space. Its deformation function is written as:
\begin{equation}
    D(\mathbf{x}) = t_{\mathbf{x}}^{c} \cdot [u, v, w]^{\top} + \left \langle \mathbf{x} - t_{\mathbf{x}}^{o} \cdot [u, v, w]^{\top}, \mathbf{n}_{t_\mathbf{x}}^{o} \right \rangle \mathbf{n}_{t_{\mathbf{x}}}^{c},
\end{equation}
where $t_{\mathbf{x}}^{c}, t_{\mathbf{x}}^{o}$ are the 3D point's nearest triangle on the canonical and observed SMPL respectively, $\mathbf{n}_{t_{\mathbf{x}}}^{c}, \mathbf{n}_{t_{\mathbf{x}}}^{o}$ are their normal values, and $[u, v, w]$ are the barycentric coordinates. \Eref{eq:triplane} can then be reformulated as:
\begin{equation}
    F((f_{xy}\circ D)(\mathbf{x}) + (f_{xz}\circ D)(\mathbf{x}) + (f_{yz}\circ D)(\mathbf{x})) \mapsto (\sigma, \mathbf{c}).
\end{equation}

\textbf{ENARF-GAN}
Following NARF~\cite{2021narf} (see \Sref{NARF}) that transforms the 3D point $\mathbf{x}$ into different parts of the local coordinate system, ENARF-GAN~\cite{noguchi2022unsupervised} applies triplane features to (1) provide features for estimating the color and density, and (2) predict the probability of a query point $\mathbf{x}$ belonging to a specific body part:
\begin{equation}
    F(\mathbf{f}) \mapsto (\sigma, \mathbf{c}), \quad \mathbf{f} = \sum_{k=1}^{K} p^k * \mathbf{f}^k,
\end{equation}
where $\mathbf{f}^k = \sum_{ij\in(xy, yz, xz)} F_{ij}(\mathbf{x}_c^k)$, $p^k=p_{xy}^kp_{xz}^kp_{yz}^k$, and $K$ is the number of body parts. After obtaining the RGB image via volume rendering, it further applies two StyleGAN2 generators separately for generating the foreground and background images, ensuring high fidelity of both the articulated human and its surrounding environment.

\textbf{HumanGen}
To achieve detailed geometry and realistic $360^{\circ}$ free-view rendering, HumanGen~\cite{jiang2022humangen} combines a 3D human reconstruction prior with a 3D-GAN network, utilizing a disentangled optimization for geometry and texture. The process begins with a pre-trained 2D generator $G_{2D}$ to generate an anchor image based on the latent code $z$. The anchor image serves a dual purpose within the framework: (1) For geometry reconstruction, a pre-trained PIFuHD~\cite{saito2020pifuhd} reconstructs the geometry based on SDF estimation using the anchor image. (2) For the texture branch, a triplane generator takes the latent code $z$ as input to generate color and blending weight; Later on, HumanGen integrates color, blending weight, and UV color information from the anchor image for each 3D query point to estimate texture values.

\textbf{EVA3D}
Improving from NARF~\cite{2021narf}, which predicts a point's probability of belonging to a specific body part, EVA3D~\cite{hong2022eva3d} takes a step further by dividing the human avatar into 16 parts based on the SMPL~\cite{SMPL:2015} model. Each part corresponds to a dedicated NeRF~\cite{mildenhall2021nerf} network, enabling localized modeling for enhanced accuracy and detail. Specifically, for each body part $k$, EVA3D applies a sub-network $F_k$ to model the local bounding box $\{\mathbf{b}_{min}^k, \mathbf{b}_{max}^k\}$. For a 3D query point $\mathbf{x}$ within the $k$-th bounding box, the density and color are predicted by:
\begin{equation}
    F_k(\mathbf{x}_k) \mapsto (\sigma^k, \mathbf{c}^k), \quad \text{where} \quad  \mathbf{x}_k = \frac{2 \mathbf{x} - (\mathbf{b}_{min}^k + \mathbf{b}_{max}^k)}{\mathbf{b}_{max}^k - \mathbf{b}_{min}^k}.
\end{equation}
If $\mathbf{x}$ falls in multiple bounding boxes, EVA3D uses a window function~\cite{lombardi2021mixture} to linearly blend the predicted values:
\begin{subequations}
    \begin{align}
        (\sigma, \mathbf{c}) &= \frac{1}{\sum \omega_k} \sum_{k \in \mathbb{K}} \{ \sigma^k, \mathbf{c}^k\},\\
        \omega_k &= \text{exp}(-m(\mathbf{x}_k(x)^n + \mathbf{x}_k(y)^n + \mathbf{x}_k(z)^n)),
    \end{align}
\end{subequations}
where $m, n$ are chosen empirically. Note that EVA3D also employs the deformation field to transform $\mathbf{x}$ from the observation space to the canonical space. Please refer to their paper for detailed illustrations.

Besides the above methods, GET3D~\cite{gao2022get3d} utilizes a differentiable explicit surface extraction method to directly optimize textured 3D meshes and a differentiable rendering technique, which can be directly used by 3D rendering engines. Next3D~\cite{sun2023next3d} advances 3D generation further by learning generative neural textures based on parametric mesh templates and mapping them onto three triplanes through rasterization to achieve both deformation accuracy and topological flexibility. 3DAvatarGAN~\cite{abdal20233davatargan}, for the first time, offers generation, editing, and animation of personalized avatars obtained from a single image via a domain-adaption framework. TriPlaneNet~\cite{bhattarai2024triplanenet} proposes to directly operate in the triplane space instead of the GAN parameter space by building upon a feed-forward convolutional encoder for the latent code and extending it with a fully convolutional predictor of triplane numerical offsets.

\begin{figure}[t]
  \centering
  \includegraphics[width=8.5cm]{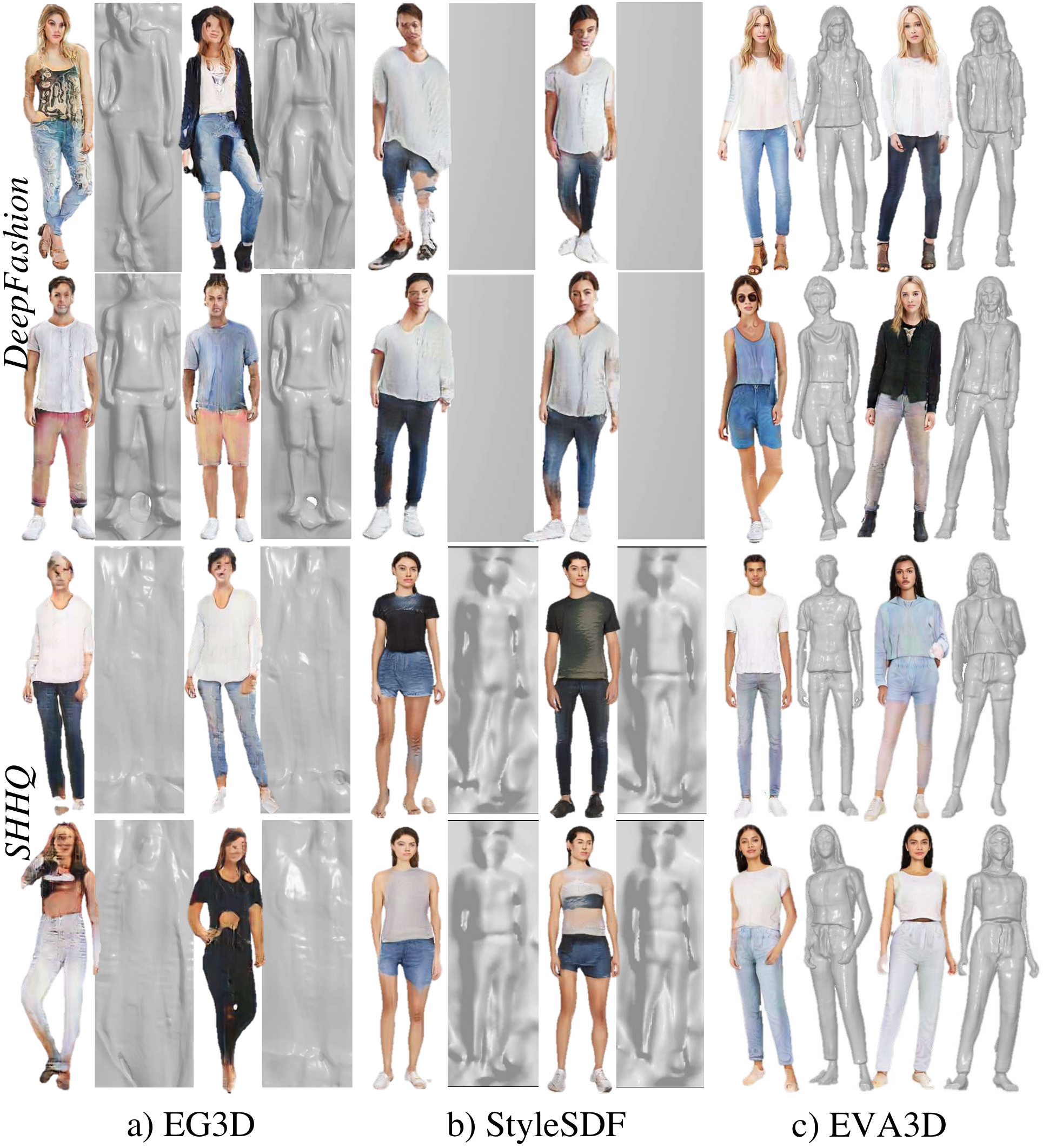}\\
  \caption{Qualitative comparison of GAN-based methods. Figure obtained from \cite{hong2022eva3d}.}
  \label{fig:gancomparison}
\end{figure}

\begin{table}[ptb]
\scriptsize
\centering
\caption{Quantitative comparison of GAN-based methods. Results obtained from~\cite{hong2022eva3d}.}
\setlength{\tabcolsep}{2pt}
\renewcommand\arraystretch{1.2}
\begin{tabular}{@{}l||c|c|c|c||c|c|c|c@{}}
\toprule
\multicolumn{1}{c||}{\multirow{2}{*}{\textbf{Methods}}}  & \multicolumn{4}{c||}{\emph{DeepFashion}~\cite{liu2016deepfashion}} & \multicolumn{4}{c}{\emph{SHHQ}~\cite{fu2022stylegan}} \\
\cline{2-9} & FID $\downarrow$ & KID $\downarrow$ & PCK $\uparrow$ & Depth $\downarrow$ & FID $\downarrow$ & KID $\downarrow$ & PCK $\uparrow$ & Depth $\downarrow$ \\ 
\midrule
EG3D~\cite{chan2022eg3d} & 26.38 & 0.014 & - & 0.0779 & 32.96 & 0.033 & - & 0.0296 \\
StyleSDF~\cite{or2022stylesdf} & 92.40 & 0.136 & - & 0.0359 & 14.12 & 0.010 & - & 0.0300 \\
EVA3D~\cite{hong2022eva3d} & 15.91 & 0.011 & 87.50 & 0.0272 & 11.99 & 0.009 & 88.95 & 0.0177 \\
\bottomrule
\end{tabular}
\label{tab:gancomparison}
\end{table}

\subsection{Discussion and Limitations of GAN-based Methods}

Upon analyzing the quantitative evaluations presented in~\tref{tab:gancomparison}, it becomes evident that incorporating triplane in recent 3D-aware GAN networks significantly enhances performance across both 2D and 3D metrics. This conclusion is further supported by the qualitative visualizations depicted in~\fref{fig:gancomparison}. Notably, even compared to StyleSDF, EG3D produces suboptimal outcomes, underscoring the importance of integrating 3D priors such as SMPL within 3D-aware GAN architectures.

Unfortunately, while 3D-aware GAN techniques achieve promising 3D human generation results by training the network only on 2D datasets, they face challenges including (1) the presence of artifacts like blurs, distortions, and noise in the generated images, and (2) a dependency on the training datasets, which limits the ability to generate content beyond the scope of the training datasets. We will then dive into more generalizable 3D human generation methods that leverage large vision-language models like Contrastive Language-Image Pretraining (CLIP)~\cite{radford2021learning} and diffusion models~\cite{stable-diffusion, saharia2022photorealistic} in the next section.

%% file: Sections/6_CLIP-based/CLIP-based.tex
\section{3D Human Generation via Large Language Models}
\label{sec:generation}
To exploit the potential of 2D generative image models such as Contrastive Language-Image Pretraining (CLIP)~\cite{radford2021learning} and diffusion model~\cite{stable-diffusion, zhang2023controlnet, liu2023zero} for generating 3D contents, methods have been proposed to optimize the 3D representation (e.g., mesh, point cloud, NeRF, 3D Gaussians) based solely on the text prompts. 

\subsection{CLIP-based 3D Human Generation}

In this subsection, we first discuss the methods that utilize the pre-trained CLIP model. Typically, CLIP is capable of mapping images and text to the same feature spaces, allowing for comparison of their similarities.

DreamField is one of the first methods that integrate 3D representations with pre-trained language models by evaluating the similarity between renderings generated from 3D contents and text prompts.
Specifically, the proposed self-optimization manner involves: (1) initializing the NeRF to a unit sphere; (2) rendering RGB images from randomly sampled camera direction during each iteration; (3) calculating the similarity distance between the rendered image and text via CLIP as the loss function; (4) backpropagating the loss to optimize the NeRF parameters. Through iterative optimization, DreamField effectively aligns NeRF with the text prompt, thus bridging the gap between natural language and 3D content generation.

Following DreamField, Text2Mesh~\cite{michel2022text2mesh} proposes to simultaneously render multiple images from various viewpoints at each iteration to enhance the performance. CLIP-Mesh~\cite{mohammad2022clip} introduces a set of render augmentations and incorporates a text-to-image embedding prior. CLIPXPlore~\cite{hu2023clipxplore} maps the encoded CLIP code to its associated shape code to ensure a coherent connection between CLIP and shape latent spaces.
Unfortunately, considering the complexity of human subjects, these methods often fall short in generating topologically and structurally correct 3D human models.

\textbf{AvatarCLIP}
To facilitate efficient 3D human generation, AvatarCLIP~\cite{hong2022avatarclip} introduces a novel approach that involves initializing the NeuS~\cite{wang2021neus} model with a predefined SMPL model, instead of the typical use of a sphere unit in CLIP-based optimization. Alongside 3D human generation, AvatarCLIP expands its method to generate motion sequences for 3D human animation based on text prompts. This process leverages a pre-calculated codebook and a pre-trained motion VAE (Variational Autoencoder) model~\cite{kingma2013auto}. By employing these resources, the pre-generated 3D human model can be animated using the SMPL-based deformation described in \Eref{eq:deformation}.

Following AvataCLIP, MotionCLIP~\cite{tevet2022motionclip} trains a transformer-based motion auto-encoder to reconstruct motion while being aligned to its text label’s position in the CLIP space.
T2M-GPT~\cite{zhang2023t2m} takes the CLIP text embedding as the language prior for text-motion model training.
AttT2M\cite{zhong2023attt2m} proposes body-part attention to learn a discrete latent space and global-local motion-text attention to learn the sentence and word level motion-text cross-modal relationship.
Wu \emph{et al.}~\cite{wu2023high} introduce the descriptive code space as an intermediary for the mapping from the text embedding space
to the 3D face parametric space.

Unfortunately, despite CLIP's advantages of avoiding expensive and hard-to-obtain 3D datasets, it still struggles with creating realistic 3D meshes and lacks broad generalization in human motion. This limitation mainly arises from CLIP's constrained capacity to fully comprehend complex human languages.

%% file: Sections/7_Diffusion_model-based/Diffusion_model-based.tex
\subsection{Diffusion Model-based 3D Human Generation and Editing}

Unlike CLIP-based~\cite{radford2021learning} methods that embed image and text into a shared latent space to compare their similarities and learn associations, diffusion models convert random Gaussian noise into structured data by a Markov process with a series of denoising steps: 
\begin{equation}
    p_\theta\left(\mathbf{x}_{0: T}\right)=p\left(\mathbf{x}_T\right) \prod_{t=1}^T p_\theta\left(\mathbf{x}_{t-1} \mid \mathbf{x}_t\right),
\end{equation}
where $p_{\theta}(\mathbf{x}_{0:T})$ is the joint distribution over all the states of the process, $p(\mathbf{x}_T)$ is the distribution at the final time step $T$, which is typically assumed to be a standard Gaussian distribution and represents the data in its most noisy form. $\prod_{t=1}^{T} p_{\theta}(\mathbf{x}_{t-1}|\mathbf{x}_t)$ is the sequential multiplication of conditional distributions, each representing the probability distribution of the state at time step $t-1$ given the state at time step $t$.

While CLIP may lead to less precise interpretations of complex instructions, diffusion models, on the other hand, bypass linguistic ambiguities inherent in language-image pairings and focus on the step-by-step refinement of visual data, and perform better in 3D human representations.

Benefiting from the development of text-guided diffusion models $\phi$~\cite{stable-diffusion, balaji2022ediffi}, DreamFusion~\cite{poole2022dreamfusion} proposes a novel Score Distillation Sampling (SDS) to enable the generation of a 3D scene $g(\theta)$.
Let's denote the RGB rendering from NeRF as $I$ and the text embedding as $y$. SDS strategy firstly involves encoding $I$ to derive the latent features $z$ and introducing random noise $\epsilon$ to $z$ to generate a noisy latent variable $z_t$. A pre-trained denoising function $\epsilon_\phi\left(z_t; y, t\right)$ is then employed to predict the added noise. The SDS loss is defined as the difference between predicted and added noise, and its gradient is given by:
\begin{equation}
    \nabla_\theta \mathcal{L}_{\mathrm{SDS}}(\phi, g(\theta))=\mathbb{E}_{t, \epsilon}\left[w(t)\left(\epsilon_\phi\left(z_t ; y, t\right)-\epsilon\right) \frac{\partial z}{\partial x} \frac{\partial x}{\partial \theta}\right],
\end{equation}
where $w(t)$ weights the loss from noise level $t$. The SDS gradients will be back-propagated to optimize $g(\theta)$, generating expressive 3D content from the text prompt.

Following DreamFusion, many other methods have been developed to improve the performance from different perspectives. Among them, Latent-NeRF~\cite{metzer2022latent-nerf} employs a latent diffusion model to optimize NeRF in the latent space, largely increasing the training efficiency. Magic3D~\cite{lin2023magic3d} utilizes a coarse-to-fine strategy that leverages latent diffusion model and \textsc{DMTet}~\cite{shen2021deep} for high-resolution 3D content generation. Fantasia3D~\cite{chen2023fantasia3d} disentangles the generation process into geometry and texture generation to enhance the performance. TEXTure~\cite{richardson2023texture} innovates in texture generation, transfer, and editing by using a pre-trained depth-to-image diffusion model and applying an iterative scheme that paints a given 3D model from different viewpoints. ProlificDreamer~\cite{wang2023prolificdreamer} introduces a Variational Score Distillation (VSD) to improve the quality and diversity of the generated 3D contents. MVDream~\cite{shi2023MVDream} fine-tunes a multi-view diffusion model to produce consistent multi-view 3D generations.
However, these methods consistently face similar limitations in generating 3D human bodies. For example, they cannot control human motions, and the results often lack hands or feet, with inconsistencies in geometry and texture.

\subsubsection{3D Human Generation}

\textbf{DreamAvatar}
DreamAvatar~\cite{cao2023dreamavatar} designs a dual-observation space in which the canonical space and observation space are jointly optimized using a shared NeRF module. The connection between these two spaces is established through a deformation field (as in \Eref{lbs-nerf}), encompassing articulated deformation and non-rigid motion. Additionally, DreamAvatar utilizes SMPL-derived density fields that allow the optimized NeRF to evolve from the density field derived from the SMPL model:
\begin{subequations}
    \begin{align}
        \bar{\sigma}_{c} &= \max(0, \mathtt{softplus}^{-1}(\frac{1}{a}\mathtt{sigmoid}(-d_c/a))), \\
        \bar{\sigma}_{o} &= \max(0, \mathtt{softplus}^{-1}(\frac{1}{a}\mathtt{sigmoid}(-d_o/a))),
    \end{align}
\end{subequations}
where $\mathtt{sigmoid}(x) = 1 / (1 + \mathrm{e}^{-x})$, $\mathtt{softplus}^{-1}(x) = \log (\mathrm{e}^x-1)$, $d_c, d_o$ are signed distance to the corresponding SMPL model, and $a$ is a predefined hyperparameter~\cite{xu2022dream3d}. The density and color values of DreamAvatar's dual-observation space are then derived by:
\begin{subequations}
    \begin{align}
        F(\mathbf{x}_c, \bar{\sigma}_{c}) &= F_{\theta}(\gamma(\mathbf{x}_c)) + (\bar{\sigma}_{c}, \mathbf{0})  \mapsto (\sigma_c, \mathbf{c}_c), \\
        F(\mathbf{x}_o, \bar{\sigma}_o) &= F_{\theta}(\gamma(\hat{\mathbf{x}}_c)) + (\bar{\sigma}_o, \mathbf{0})  \mapsto (\sigma_o, \mathbf{c}_o),
    \end{align}
\end{subequations}
where $\hat{\mathbf{x}}_c$ is $\mathbf{x}_o$'s corresponding canonical point.
DreamAvatar efficiently enables the distillation of the well-optimized texture and geometry from the canonical space to the observation space, achieving high-quality and controllable avatar generation under user-guided human pose. 

\textbf{AvatarCraft}
Following DreamFusion and AvatarCLIP, AvatarCraft~\cite{jiang2023avatarcraft} leverages NeuS~\cite{wang2021neus} (initialized from SMPL) and a pre-trained diffusion model to generate 3D human avatars. Specifically, AvatarCraft proposes to divide the canonical avatar into face and body bounding boxes according to the SMPL model and separately render them in a coarse-to-fine manner to recover higher-resolution texture and geometry. Since the optimization process closely aligns with the SMPL model, the generated human models can be readily animated using \Eref{eq:deformation}.

\begin{figure*}[t]
  \centering
  \includegraphics[width=\textwidth]{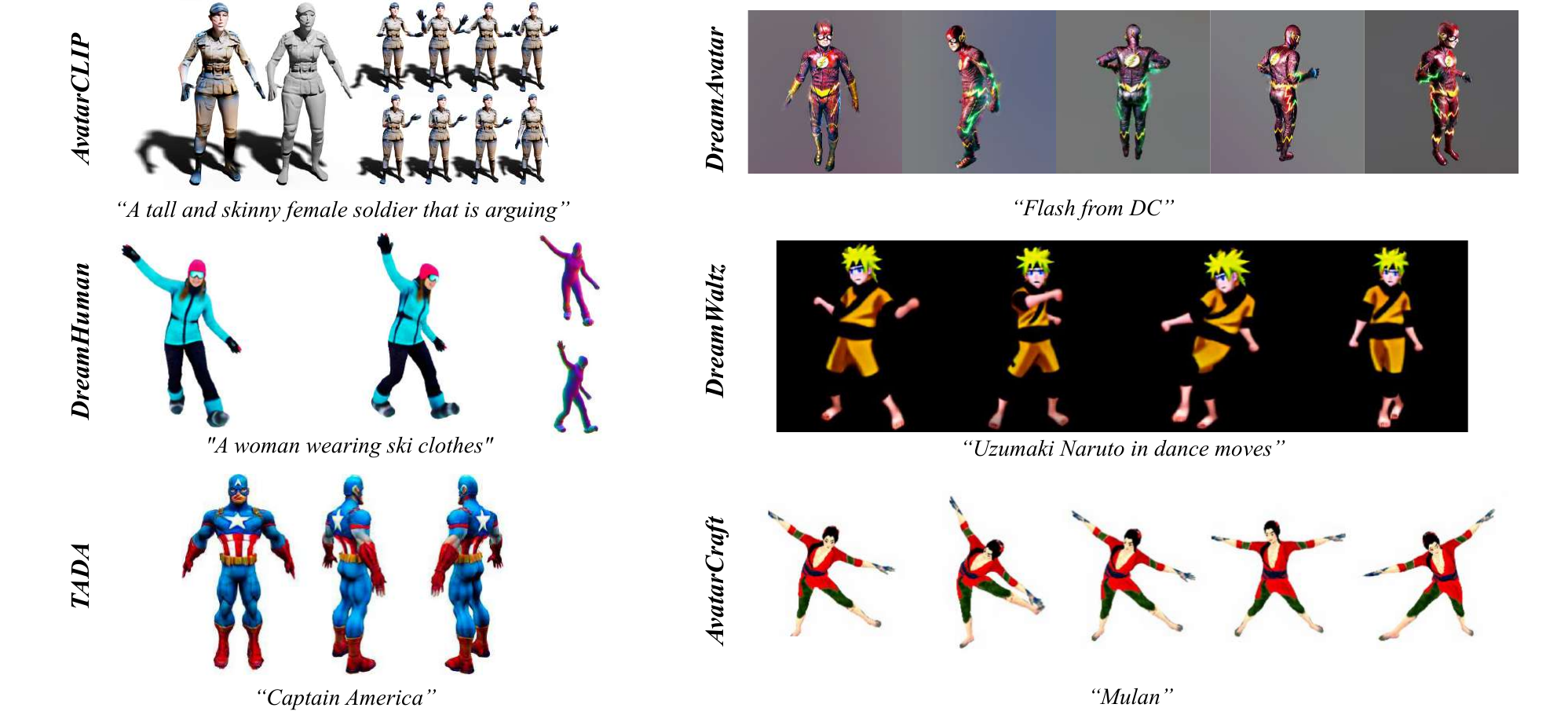}\\
  \caption{Results of various 3D human generation methods. Figure obtained from \cite{hong2022avatarclip, cao2023dreamavatar, jiang2023avatarcraft, huang2023dreamwaltz, kolotouros2023dreamhuman, liao2023tada}.}
  \label{fig:diffusion1}
\end{figure*}

\textbf{DreamWaltz}
Instead of initializing the optimization process from SMPL or constraining it to SMPL surfaces, DreamWaltz~\cite{huang2023dreamwaltz} introduces an alternative approach by replacing the Stable Diffusion model~\cite{stable-diffusion} with a pose-conditioned ControlNet~\cite{zhang2023controlnet}. In addition to rendering RGB images, the network simultaneously generates pose skeleton images from the SMPL model at each iteration. These skeleton images, along with RGB renderings and text embeddings, are input into ControlNet, ensuring consistent 3D gradients for the SDS loss. Its further proposes occlusion culling~\cite{pantazopoulos2002occlusion} for the skeleton images to remove invisible parts, addressing the multi-face ``Janus'' problem.

\textbf{DreamHuman}
In contrast, DreamHuman~\cite{kolotouros2023dreamhuman} integrates imGHUM~\cite{alldieck2021imghum} as its deformable NeRF to enforce constraints on the human body. DreamHuman introduces a semantic zooming strategy to enhance the generation quality. Specifically, during each iteration, the network identifies different body parts (e.g., hands, head, arms, legs, etc.) and performs zoomed-in rendering separately for each part to calculate the SDS. The renderings capture intricate details and therefore help refine the texture and geometry.

\textbf{ZeroAvatar}
Unlike other methods that utilize text prompts to generate 3D avatars, ZeroAvatar~\cite{weng2023zeroavatar} takes a single human image as input. Given a single image, ZeroAvatar (1) estimates the body pose, shape, and UV map to initialize the density field, (2) optimizes the geometry via SDS loss, integrating depth from the posed body model for enhanced accuracy, and (3) applies the inferred UVs to parts of the body that are not visible in the original image for a complete appearance. Specifically, for texture completion, it first uses DensePose~\cite{Guler2018DensePose} to regress the UV coordinates from the image, and samples RGB colors to fill in the visible region of the UV map. The symmetrical areas of the visible region are then predicted and used as a prior during optimization, serving as a stronger guide than SDS.

\textbf{TADA}
TADA~\cite{liao2023tada} adopts 3D mesh representations and directly initializes it from SMPL-X~\cite{SMPL-X:2019} surfaces, achieving robust 3D human generation and animation. After initializing from the SMPL-X surface with uniformly sampled points, TADA learns the SMPL-X parameters $\beta$, $\theta$, $\phi$ and a displacement $\mathbf{D}$ which accounts for personalized details that are independent of pose, shape, and expression. Hence, the learnable 3D mesh can be written based on \Eref{smplx}:
\begin{subequations} \label{tada}
\begin{align}
    M(\beta, \theta, \phi, \mathbf{D})&=\mathtt{lbs}(\hat{T}(\beta, \theta, \phi, \mathbf{D}), J(\beta), \theta, \mathcal{W}), \\ 
    \hat{T}(\beta, \theta, \phi, \mathbf{D})&=\mathcal{S}(T(\beta, \theta, \phi)) + \mathbf{D},
\end{align}
\end{subequations}
where $\mathcal{S}(\cdot)$ is the mesh subdivision operation to add more details. To better align the geometry and texture for animation, it proposes to compute an additional SDS on the interpolation between normal and color image latents.

\textbf{AvatarBooth}
To generate personalized 3D avatars, AvatarBooth~\cite{zeng2023avatarbooth} introduces dual latent diffusion models to supervise the face and body generation separately.
Specifically, AvatarBooth needs a set of personalized images as input, and it first separates the input images into full body shots and headshots for fine-tuning two diffusion models. For the headshots, the pose-consistent constraint proposed in ControlNet~\cite{zhang2023controlnet} is incorporated to ensure that the generated facial images maintain consistent identity across multiple views.

\textbf{HumanNorm}
Considering that previous text-to-image diffusion models lack understanding of 3D structures, HumanNorm~\cite{huang2023humannorm} fine-tunes three diffusion models, i.e., text-to-normal diffusion model, text-to-depth diffusion model, and normal-aligned diffusion model, and adopts a disentangled optimization of the geometry and texture. It largely enhances the 2D perception of the 3D geometry while ensuring the consistency between generated geometry and texture.

We provide several visualizations in \fref{fig:diffusion1} to offer further insights into the metrics employed by different methods for generating controllable 3D human avatars.

\subsubsection{3D Human Editing}

\textbf{Control4D}
Building upon Tensor4D~\cite{shao2023tensor4d}, Control4D~\cite{shao2023control4d} introduces a 4D GAN architecture that can be edited via a ControlNet-based~\cite{zhang2023controlnet} diffusion model, showcasing high-fidelity and consistent 4D editing based on 4D data and text prompts. It first employs Tensor4D to train the implicit representation of a 4D portrait scene from the 4D data, which is then rendered into latent features and RGB images via voxel rendering, serving as inputs to the generator. Meanwhile, the ControlNet takes original images and text prompts as inputs to produce edited images. The edited images serve as ``real images'' and the generator's outputs are ``fake images'', allowing for iterative 4D editing.

\textbf{HeadSculpt}
HeadSculpt~\cite{han2023headsculpt} adopts a coarse-to-fine strategy to generate high-resolution head avatars and perform fine-grained editing solely based on text prompts. 
To address the multi-face "Janus problem" in head generation, HeadSculpt proposes Prior-driven Score Distillation (PSD) that integrates 3D head priors and view-dependent textual inversion into the diffusion model. HeadSculpt further introduces Identity-aware Editing Score Distillation (IESD) that respects optimization gradients from both the original identity and instructive prompts, achieving fine-grained editing while maintaining its identity. 

\textbf{HeadArtist}
To address the inherent limitations of SDS (i.e., over-saturation and over-smoothing), HeadArtist~\cite{liu2023headartist} optimizes a parameterized 3D head model $R(\theta)$ under the supervision of the prior distillation itself, which is called Self Score Distillation (SSD):
\begin{equation}
    \begin{array}{r}
    \nabla_\theta \mathcal{L}_{\mathrm{SSD}}=\mathbb{E}_{t, \epsilon}\left[\omega ( t ) \left(\epsilon_\pi\left(x_t ; y, t, c_L\right)-\right.\right. \\
    \left.\left.\hat{\epsilon}_\pi\left(x_t ; y, t, c_L\right)\right) \frac{\partial x}{\partial \theta}\right],
\end{array}
\end{equation}
where $c_L$ is the landmark that contains facial structure priors, $\epsilon_\pi$ and $\hat{\epsilon}_\pi$ are two pre-trained ControlNets with the same parameters, $x_t$ is the marginal distribution of the ControlNet given the text and landmarks.

\textbf{AvatarStudio}
Evolving from Instruct-NeRF2NeRF~\cite{haque2023instructnerf2nerf}, AvatarStudio~\cite{pan2023avatarstudio} achieves text-driven editing of dynamic head avatars by (1) fine-tuning pre-trained diffusion models on images with various viewpoints and time stamps, and (2) a novel view- and time-aware score distillation sampling (VT-SDS):
\begin{equation}
    \nabla_x \mathcal{L}_{VT-SDS}=w(t)\left(\epsilon_t-\Psi\left(x_t, t, \mathbf{s}, \mathbf{s}_i\right)\right),
\end{equation}
where $\mathbf{s}$ is the text embedding, $\mathbf{s}_i=\Gamma(\mathbf{P}_i)$ is a conditioning vector with $\mathbf{P}_i$ being the label assigned to the input image, and $\Psi\left(\mathrm{x}_t, t, \mathbf{s}, \mathbf{s}_i\right)$ is the noise predicted by the diffusion model.

Besides, other works also achieve good results in 3D object and human editing. Progressive3D~\cite{cheng2023progressive3d} decomposes the generation into a series of locally progressive editing steps to create precise 3D content for complex prompts. Vox-E~\cite{sella2023vox} introduces a novel volumetric regularization loss that operates directly in 3D space to maintain global coherence between the original and edited object. RODIN~\cite{wang2023rodin} trains an image encoder to extract a semantic latent vector as the conditional input of the diffusion model, allowing semantic editing of generated results. DiffusionRig~\cite{ding2023diffusionrig} learns generic and person-specific facial priors from extensive and individual datasets, respectively, for high-fidelity editing. TECA~\cite{zhang2023text} combines traditional 3D mesh models for the head, face, and upper body with NeRF for the modeling and editing of hair, clothing, and accessories.

\begin{figure}[t]
  \centering
  \includegraphics[width=8.5cm]{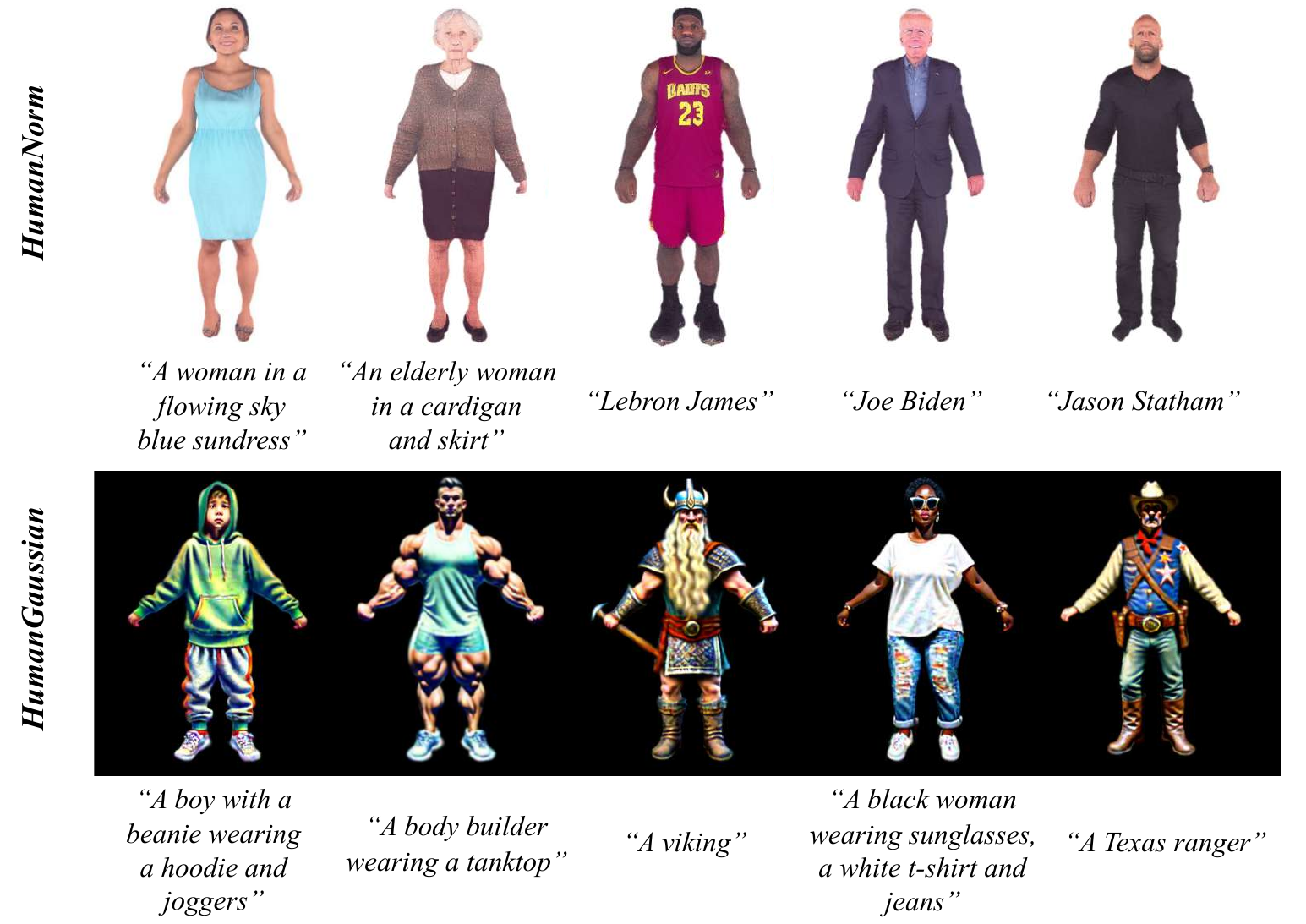}\\
  \caption{Results of HumanNorm and HumanGaussian. Figure obtained from \cite{huang2023humannorm, liu2023humangaussian}.}
  \label{fig:diffusion2}
\end{figure}
\subsection{3D Gaussian-based 3D Human Generation and Editing}
With the development of 3D Gaussian Splatting (3DGS), methods have been introduced to apply 3DGS for high-quality 3D generation with increased training efficiency.

Among them, DreamGaussian~\cite{tang2023dreamgaussian} designs a mesh extraction algorithm from 3D Gaussians and a UV-space texture refinement for both efficiency and quality. GSGEN~\cite{chen2023text} adopts a coarse-to-fine strategy for more delicate details and accurate geometry. GaussianDreamer~\cite{yi2023gaussiandreamer} utilizes a 3D diffusion model to provide priors for initialization and a 2D diffusion model to enrich the geometry and appearance. LucidDreamer~\cite{chung2023luciddreamer} leverages Stable Diffusion~\cite{stable-diffusion}, depth estimation, and explicit 3D representation for a domain-free high-quality 3D scene generation. GALA3D~\cite{zhou2024gala3d} introduces a layout-guided Gaussian model and a compositional optimization mechanism to ensure geometry and texture consistency and accurate object interactions.

\subsubsection{3D Human Generation}

\textbf{HumanGaussian}
HumanGaussian~\cite{liu2023humangaussian} is an efficient and effective framework that combines Structure-Aware SDS and Annealed Negative Prompt Guidance for high-quality 3D human generation. In Structure-Aware SDS, HumanGaussian (1) utilizes SMPL-X~\cite{SMPL-X:2019} as a prior to densely sample Gaussians on the human mesh surface, (2) train a Texture-Structure Joint Model to simultaneously denoise the image and depth conditioned on the posed skeleton, and (3) design a dual-branch SDS to jointly optimize the appearance and geometry. In Annealed Negative Prompt Guidance, HumanGaussian uses the cleaner classifier score with an annealed negative score to regularize the stochastic SDS gradient of high variance. The floating artifacts are further eliminated based on Gaussian size in a prune-only phase to enhance generation smoothness.

Different from methods~\cite{hong2022avatarclip, kolotouros2023dreamhuman, cao2023dreamavatar, huang2023dreamwaltz, liao2023tada, jiang2023avatarcraft} which aim to control the generation of avatars in complex poses (see~\fref{fig:diffusion1}), HumanNorm and HumanGaussian focus primarily on pre-training diffusion models to enhance the alignment between geometry and texture. The generated results of their approach can be observed in~\fref{fig:diffusion2}.

\textbf{HeadStudio}
HeadStudio~\cite{zhou2024headstudio} achieves high-quality and animated head avatars differently by integrating FLAME~\cite{li2017flame} into 3D Gaussian splatting and SDS. It first proposes to deform 3D Gaussian points with facial expressions via FLAME-based 3D Gaussian Splatting (F-3DGS), where each 3D point is linked to a FLAME mesh and then rotated, scaled, and translated by the mesh deformation. Subsequently, it utilizes FLAME-based Score Distillation Sampling (F-SDS) which employs FLAME-based fine-grained control signals to guide the score distillation process. Finally, it enhances generation quality by applying uniform super-resolution and mesh regularization in F-3DGS.

\begin{figure}[t]
  \centering
  \includegraphics[width=8.5cm]{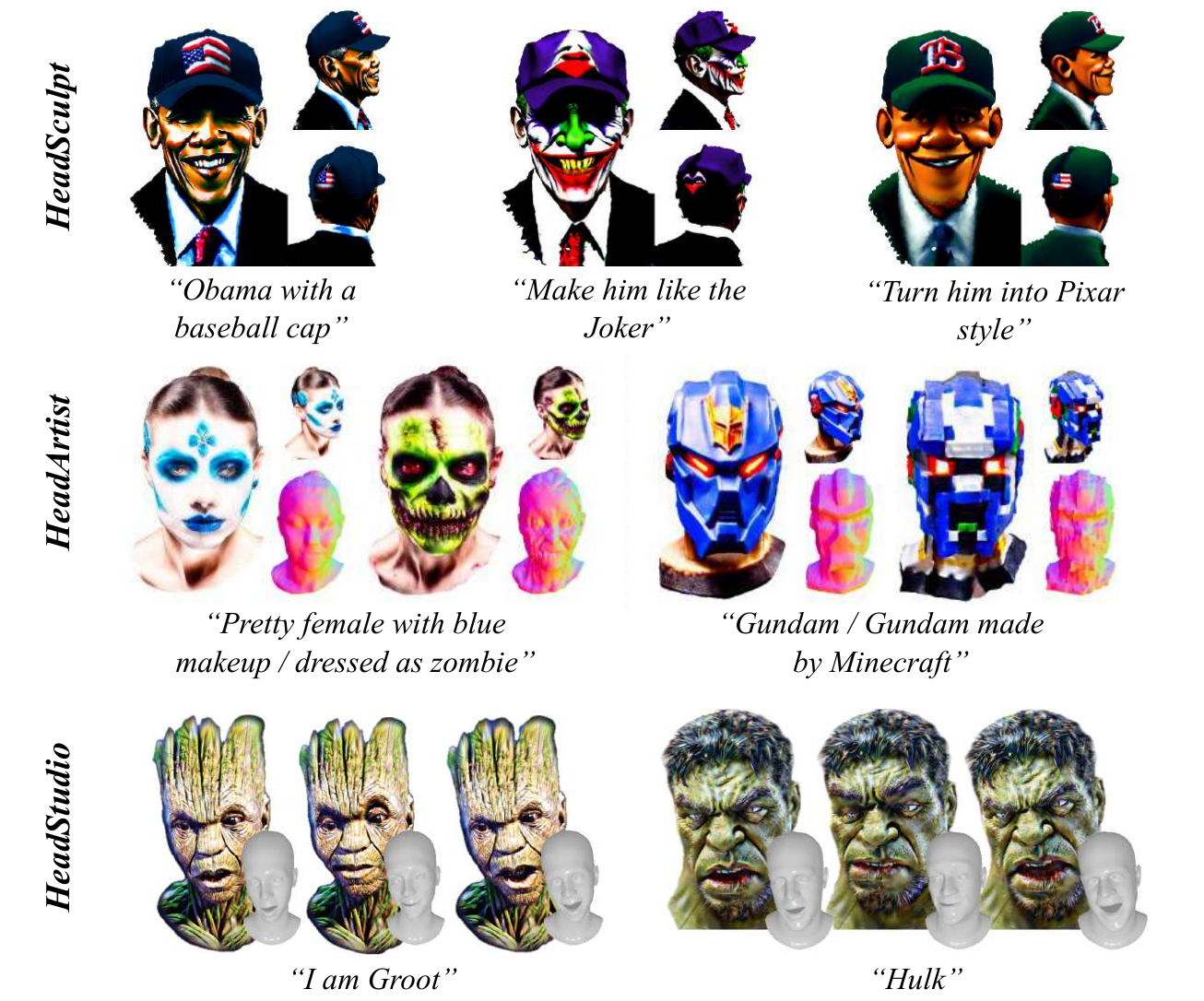}\\
  \caption{Generation or editing results of methods specifically designed for head avatars. Figure obtained from \cite{han2023headsculpt, liu2023headartist, zhou2024headstudio}.}
  \label{fig:head}
\end{figure}

We show several methods in~\fref{fig:head} that are specifically designed for head avatars, which can efficiently generate or edit human heads.

\subsubsection{3D Human Editing}
3D editing methods based on implicit 3D representations like NeRF~\cite{mildenhall2021nerf} suffer from slow processing speeds and limited control over complex scenes, while 3D Gaussian Splatting~\cite{kerbl20233d} offers an opportunity to efficiently locate user-guided semantics with enhanced efficiency.

\textbf{GaussianEditor}
GaussianEditor~\cite{chen2023gaussianeditor} achieves precise local editing via Hierarchical Gaussian Splatting (HGS) and Gaussian Semantic Tracing. By providing a 2D segmentation mask with the editing area, Gaussian Semantic Tracing first identifies the corresponding parts in the 3D scene via back-projection, followed by assigning semantic tags to the affiliated Gaussians in these areas. HGS then records the semantic attributes of all the initialized Gaussians. By inheriting this attribute to child Gaussians during the densification and pruning, HGS enables more detailed and effective local editing. Besides precise local editing, GaussianEditor can also remove objects and integrate new objects via 3D impainting: 1) When an object needs to be removed, it identifies and isolates the object using Gaussian Semantic Tracing. 2) When an object needs to be added, it creates a 3D representation of the new object, which is then converted into the Gaussians that are compatible with the HGS system. 

\begin{figure}[t]
  \centering
  \includegraphics[width=8.5cm]{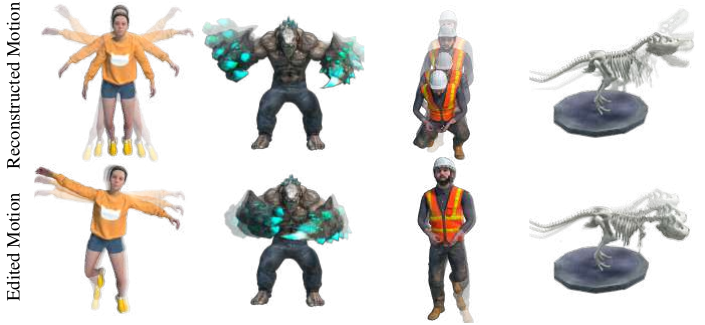}\\
  \caption{Editing results of SC-GS. Figure obtained from \cite{huang2023sc}.}
  \label{fig:sc-gs}
\end{figure}

\textbf{SC-GS}
Besides reconstruction illustrated in Section~\ref{3D Gaussian reconstruction}, SC-GS~\cite{huang2023sc} also allows for efficient motion editing by manipulating the learned control points. For each control point $p_j$, SC-GS calculates its trajectory $p_i^{\text{traj}}$ that includes its locations across $N_t(=8)$ randomly sampled time steps as:
\begin{equation}
    p_i^{\text{traj}}=\frac{1}{N_t} p_i^{t_1} \oplus p_i^{t_2} \oplus \cdots \oplus p_i^{t_{N_t}},
\end{equation}
where $\oplus$ denotes the vector concatenation operation. With the trained control points and the deformation MLP, SC-GS constructs a control point graph $\mathcal{P}^{\prime}$ that connects control points based on their trajectories. An ARAP~\cite{sorkine2007rigid} deformation module is applied to the control graph to maintain local rigidity. Specifically, given a set of user-defined points $\left\{h_l \in \mathbb{R}^3 \mid l \in \mathcal{H} \subset\right.$ $\left.\left\{1,2, \cdots, N_p\right\}\right\}$ ($N_p$ is the number of the control points), the control graph $\mathcal{P}^{\prime}$ is deformed by minimizing the ARAP energy, which can be formulated as:
\begin{equation}
    E\left(\mathcal{P}^{\prime}\right)=\sum_{i=1}^{N_p} \sum_{j \in \mathcal{N}_i} w_{i j}\left\|\left(p_i^{\prime}-p_j^{\prime}\right)-\hat{R}_i\left(p_i-p_j\right)\right\|^2,
\end{equation}
with fixed position condition $p_l^{\prime}=h_l$ for $l \in \mathcal{H}$. $w_{ij}$ is the interpolation weights for control point $p_j$ and Gaussian $G_i$. $\hat{R}_i$ is the rigid local rotation defined on each control point. Examples of editing results are shown in~\fref{fig:sc-gs}.

\textbf{TIP-Editor}
To compensate for the lack of accurate control over the specified appearance and location of the editing result in existing methods, TIP-Editor~\cite{zhuang2024tip} uses a 3D bounding box to specify the editing region for 3D scene editing. It begins with a 2D stepwise personalization that incorporates (1) a scene personalization step with a localization loss to enhance the interaction between the existing 3D scene and the edited content, and (2) a content personalization step utilizing LoRA~\cite{hu2021lora}. TIP-Editor proposes a coarse editing stage via SDS, and a pixel-level reconstruction loss to refine the texture of the 3D scene following SDEdit~\cite{meng2021sdedit}.

\begin{figure}[t]
  \centering
  \includegraphics[width=8.5cm]{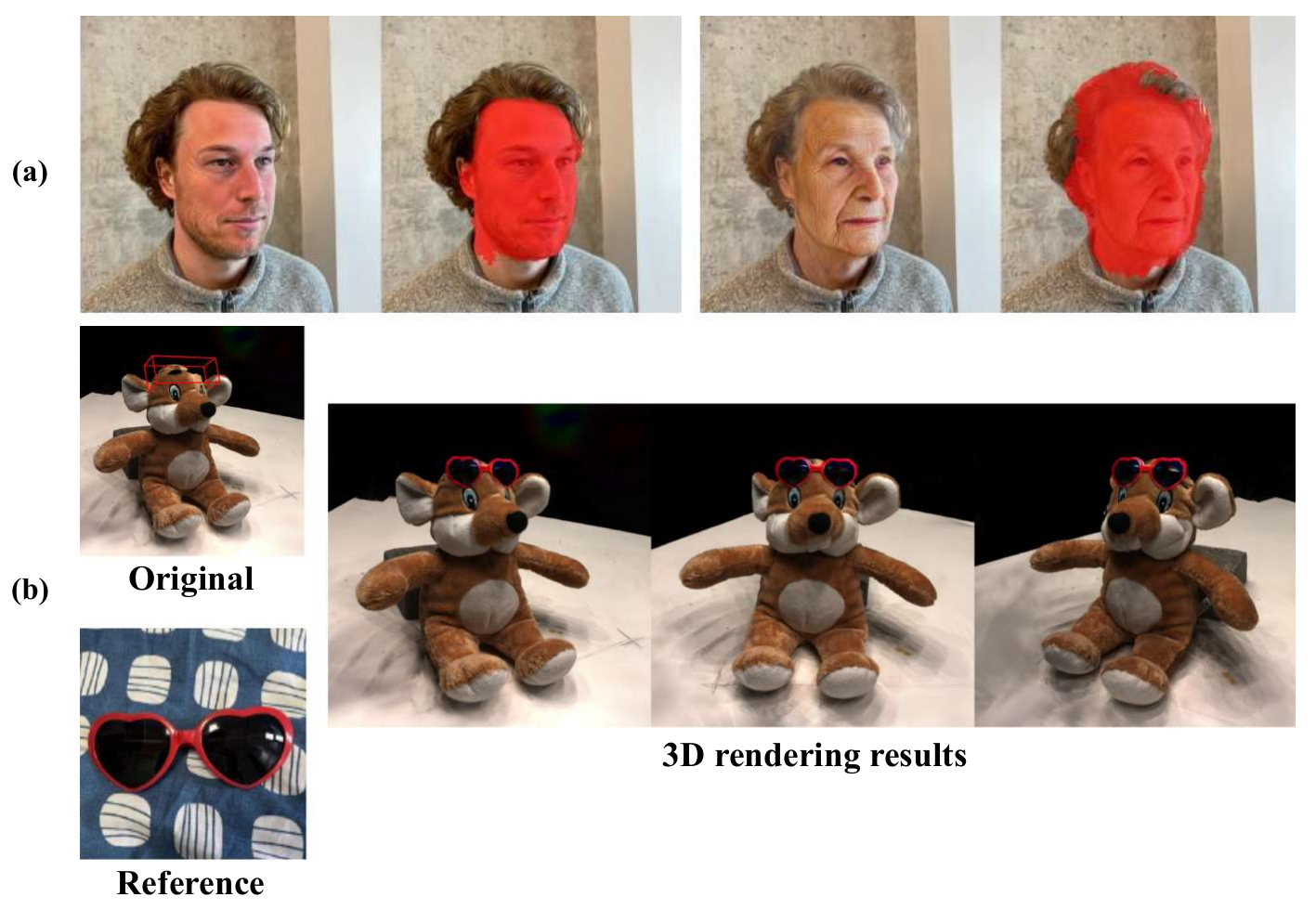}\\
  \caption{Comparison of (a) GuassianEditor and (b) TIP-Editor. Figures obtained from \cite{chen2023gaussianeditor, zhuang2024tip}.}
  \label{fig:gaussian-editing}
\end{figure}

We present several visualizations in~\fref{fig:gaussian-editing} to demonstrate the distinct metrics utilized by GaussianEditor and TIP-Editor for achieving locally controllable editing.

In addition to the above methods, 
Point’n Move~\cite{huang2023point} devises a two-stage self-prompting mask propagation process that produces 3D semantic segmentation masks from 2D image prompts for interactive scene object editing. Texture-GS~\cite{xu2024texture} separates appearance from geometry by mapping 2D texture onto the 3D surface. GaussCtrl~\cite{wu2024gaussctrl} proposes depth-conditioned editing and attention-based latent code alignment for multi-view consistent editing. VcEdit~\cite{wang2024view} also aims to maintain consistency via a Cross-attention Consistency Module and
an Editing Consistency Module.

%% file: Sections/8_Reflection/Reflection.tex
\section{Reflection}
\label{sec:reflection}
Despite recent advancements in 3D human reconstruction and generation, there are still many areas in existing methods that demand further study. In this section, we explore some prominent perspectives with the aim of offering inspiration for advancing the field. While nowadays ``reconstruction'' and ``generation'' are always integrated to achieve good results, we categorize the discussion into optimization-based techniques and feed-forward pipelines.

\subsection{Optimization-based Methods}

\subsubsection{Integration of Image References and Text Prompts}

As previously discussed, we can perceive NeRF~\cite{mildenhall2021nerf} / 3DGS~\cite{kerbl20233d} and DreamFusion~\cite{poole2022dreamfusion} as two extremes. NeRF and 3DGS rely on a multitude of images from diverse viewpoints to reconstruct the 3D scene, while DreamFusion~\cite{poole2022dreamfusion} bridges the connection between images and text prompts via pre-trained large language models and generates 3D content solely from text prompts. Consequently, NeRF and 3DGS excel in robust 3D scene reconstruction but are limited in their ability to generalize beyond trained specific subjects. On the other hand, DreamFusion and subsequent methods exhibit promising potential in generating previously unseen content distributions, but they face challenges related to quality, training efficiency, and robustness.
Therefore, we aim to explore opportunities for integrating the advantages of both image references and text prompts.

We divide the discussion into three parts based on the input type and number.

\textbf{(1) Text prompt only.} Although AvatarCLIP~\cite{hong2022avatarclip}, DreamAvatar~\cite{cao2023dreamavatar}, AvatarCraft~\cite{jiang2023avatarcraft}, DreamWaltz~\cite{huang2023dreamwaltz}, TADA~\cite{liao2023tada}, and following works~\cite{huang2023humannorm, liu2023humangaussian, zeng2023avatarbooth} introduce several metrics to control the 3D avatars and ensure structural and topological accuracy, there remain challenges in: 

\emph{(I) Training efficiency and stability.} Existing methods always require hours to optimize a single 3D model, which is impractical for real-world applications. One direction to improve the situation would be reducing the stochastic feature of the SDS. Methods like ProlificDreamer~\cite{wang2023prolificdreamer}, Consistent3D~\cite{wu2024consistent3d}, and SteinDreamer~\cite{wang2023steindreamer} seek to minimize the variations of SDS, enhancing optimization robustness but not fully addressing training deficiency and stability issues. With the release of 3D large datasets, e.g, Objaverse~\cite{deitke2023objaverse}, Objaverse-XL~\cite{deitke2024objaverse}, and OmniObject3D~\cite{wu2023omniobject3d}, recent methods~\cite{shi2023MVDream, shi2023zero123++} have proposed to fine-tune the diffusion model for generating multi-view consistent images. Unfortunately, initialization is still the key factor that constrains their training efficiency and stability. Another direction for obvious improvement would be the introduction of retrieval techniques that can retrieve the related model from 3D datasets based on text prompts, and the retrieved 3D model will serve as the basis for further optimization. 

\emph{(II) Geometric quality.} Since SDS is more stochastic than reconstruction-based loss functions, existing methods always exhibit noisy geometry. The problem becomes more severe in Gaussian-based generation, as SDS gradients will directly impact the Gaussian position and attributes instead of the MLP in NeRF. One may consider improving the 3D representation based on existing approaches like Mosaic-SDF~\cite{yariv2023mosaic}, FlexiCube~\cite{shen2023flexible}, Hierarchical Gaussian Splatting~\cite{chen2023gaussianeditor}, and SuGaR~\cite{guedon2023sugar}. We believe the disentanglement of the position and texture attributes of 3D Gaussian Splatting will also provide a promising pathway. 

\emph{(III) Reliance on SMPL prior.} As discussed before, current methods rely heavily on SMPL~\cite{SMPL:2015} and SMPL-X~\cite{SMPL-X:2019}, which, despite providing useful 3D priors, bring certain limitations. Specifically, the inherent characteristics of SMPL, e.g., containing only minimal clothing topology, largely constrain avatar customization, especially for the generation and animation of loose clothing. Learning from PIFu~\cite{saito2019pifu}, an alternative approach could involve pre-training an encoder that can perceive depth, silhouette, and semantic information.

\textbf{(2) Single image and text prompt.} With only a single image as input, the situation is similar to the single-view reconstruction in PIFu and related works, where the limited information in only one image poses issues, especially for human subjects where self-occlusion and depth-ambiguity often occur. To mitigate this, incorporating text prompts and pre-trained diffusion models could provide supplementary guidance for unseen views. In 3D object generation, methods like Zero-1-to-3~\cite{liu2023zero}, One-2-3-45~\cite{liu2023one}, Magic123~\cite{qian2023magic123}, and others~\cite{watson2022novel, zhou2023sparsefusion, chan2023genvs, liu2023one++, melas2023realfusion, seo2023let, lin2023consistent123, ye2023consistent, liu2023syncdreamer} estimate the image from other camera views via diffusion model, creating more robust 3D models from a single image and text prompts. TeCH~\cite{huang2023tech} proposes to apply the garment parsing model (SegFormer~\cite{xie2021segformer}) and BLIP~\cite{li2022blip} to extract text prompts from the input image for further optimization. However, capturing all visual attributes to accurately reconstruct unseen areas and obtaining the correct pose remains a challenge. Thus, exploring better uses of diffusion models, e.g., adding more conditions and leveraging geometric correspondences, would be a promising future direction.

\textbf{(3) Multi-view images and text prompt.} Multi-view images can provide more information in the 3D space but demand efficient feature fusion methods, a long-standing problem in 3D human reconstruction. Meanwhile, ensuring consistency between multi-view images and text prompts remains an open challenge. With the development of diffusion models, Guide3D~\cite{cao2023guide3d} generates multi-view images via ControlNet~\cite{zhang2023controlnet} and textual inversion~\cite{gal2022textual-inversion}, while facing issues like inconsistent poses, texture, and orientations. Although Guide3D introduces a joint optimization of multi-resolution \textsc{DMTet}~\cite{shen2021deep} grids, the geometric quality still lacks essential details. MVDream~\cite{shi2023MVDream} proposes to train a diffusion model for generating multi-view images from text prompts but still falls short with human subjects. Moreover, both Guide3D and MVDream deal with sparse-view images. Thus, designing a diffusion model that is capable of generating consistent and dense multi-view images needs to be studied. We believe recent video diffusion models~\cite{blattmann2023stable, bar2024lumiere, guo2023i2v, gong2024atomovideo} would provide inspiration for better spatial and temporal consistency.

\subsubsection{3D Human Editing}
Given user preferences for privacy in virtual world applications like VR / AR, 3D human editing becomes a useful area of research. Existing methods for 3D human editing mainly detail under a global style, which are effective strategies to apply diffusion models but inevitably change remaining parts or the environment. TIP-Editor~\cite{zhuang2024tip} achieves local editing via the usage of a pre-defined bounding box. Yet, achieving automatic local editing represents a more intuitive and effective future direction.

\subsubsection{3D Human Animation}

While 3D human generation can yield attractive results, the static 3D model cannot be readily applied in films, gaming, etc. Therefore, generating realistic human motion sequences and animating 3D human avatars based on large language models are promising research directions. Existing forms of human-motion generation~\cite{tevet2022human, zhang2024motiondiffuse, yuan2023physdiff, zhang2023remodiffuse, petrovich2023tmr, jin2024act, zhu2023human} are all effective ways to make diffusion models more controllable in generating motion sequences from text prompts. However, these methods still fall short of real-world demands and realism. Meanwhile, generating human-object interaction motion sequences remains a hard open problem. On the other hand, by leveraging video diffusion models, DreamGaussian4D~\cite{ren2023dreamgaussian4d} and 4D-fy~\cite{bahmani20234d} attempt multi-stage animations. However, their scale of animation is still minimal, and they struggle to represent complex poses or movements accurately. Therefore, exploring user-guided and controllable control points as the anchor would be an encouraging direction in addressing these problems.

\subsection{Feed-forward Methods}
Feed-forward methods aim to pre-train a model on 3D datasets, offering text-based or image-based inference with much less time compared with optimization-based methods. While only a limited number of recent feed-forward methods specifically cater for human subjects, this section focuses on discussing existing methods and assessing their potential for 3D human modeling.

\subsubsection{3D Foundation Model}

Recently, the advent of extensive 3D datasets such as Objaverse~\cite{deitke2023objaverse, deitke2024objaverse} and OmniObject3D~\cite{wu2023omniobject3d} has fueled progress in 3D modeling techniques. Point-E~\cite{nichol2022point-e} and Shap-E~\cite{jun2023shap} train a 3D foundation model which can generate text-guided point clouds within minutes. MVDream~\cite{shi2023MVDream}, Zero-1-to-3~\cite{liu2023zero}, and Zero123++\cite{shi2023zero123++} put forth the idea of training diffusion models for generating consistent multi-view images using either a text prompt or a single image as input. Other methods, including SyncDreamer\cite{liu2023syncdreamer}, Wonder3D~\cite{long2023wonder3d}, One-2-3-45~\cite{liu2023one}, UniDream~\cite{liu2023unidream}, HexaGen3D~\cite{mercier2024hexagen3d}, Sculpt3D~\cite{chen2024sculpt3d}, MVDiffusion++\cite{tang2024mvdiffusion++}, Make-Your-3D\cite{liu2024make}, and more~\cite{li2023sweetdreamer, liu2023one++, wang2023imagedream, qian2023magic123, lorraine2023att3d, kwak2023vivid, shi2023toss}, focus on achieving 3D-consistent generation with intricate details in geometry and texture from generated multi-view images. Additionally, the advancements in video diffusion models have led to the development of V3D~\cite{chen2024v3d}, which leverages the temporal consistency characteristic of such models and fine-tunes pre-trained stable video diffusion models to generate dense multi-view images from single-view inputs. Yet, building a 3D feature space from the multi-view generated images to enhance the level of detail remains an unresolved challenge. Furthermore, accurately capturing the pose and clothing topologies of humans poses challenges due to the intricate nature of the human body. Hence, developing robust algorithms that can deal with both general objects and human subjects becomes crucial.

\subsubsection{Large Reconstruction Model}\label{lrm}

Building on the success of transformers~\cite{vaswani2017attention} and DINO~\cite{caron2021emerging, oquab2023dinov2}, LRM (Large Reconstruction Model)~\cite{hong2023lrm} proposes a large transformer-based architecture to decode 3D triplane representation from DINO-encoded image features. Taking a single image as input, LRM reduces 3D modeling time to about 5 seconds. Following LRM, PF-LRM~\cite{wang2023pf} achieves joint pose and shape prediction in 3D object reconstruction from a few unposed images. DMV-3D~\cite{xu2023dmv3d} also achieves fast 3D generation from text or a single image via a multi-view 2D image diffusion model and an LRM-based multi-view denoiser that reconstructs noise-free triplane NeRFs. Instant3D~\cite{li2023instant3d} employs a two-stage approach that generates multi-view images from text prompts and reconstructs the 3D model via the large transformer-based module. More recently, TripoSR~\cite{tochilkin2024triposr} enhances network quality and efficiency, and 3DTopia~\cite{hong20243dtopia} introduces hybrid diffusion priors that include both text-conditioned triplane latent diffusion model and 2D diffusion priors to generate high-quality 3D objects. LGM~\cite{tang2024lgm}, CRM~\cite{wang2024crm}, and GRM~\cite{xu2024grm} explore various 3D representations, e.g., 3DGS~\cite{kerbl20233d}, FlexiCube~\cite{shen2023flexible}, for improved performance. However, similar problems in capturing the details and dynamics of human avatars as 3D foundation models can be observed. Although HumanLRM~\cite{weng2024single} proposes to distill multi-view reconstruction into single-view via a conditional triplane diffusion model for human subjects, its performance does not surpass traditional 3D implicit function-based human reconstruction methods.

\subsubsection{Combination of Explicit Representaion and 3D Gaussian Splatting}\label{triplane}

While 3DGS~\cite{kerbl20233d} has demonstrated notable performance, it possesses a non-structural nature. The optimization process prioritizes updating the appearance of 3D Gaussians rather than directly moving them to the desired 3D locations. To enhance geometric optimization, researchers have been exploring methods that separately optimize the geometry (i.e., 3D point positions) and Gaussian attributes (texture). One such approach is Triplane-Meets-Gaussian-Splatting~\cite{zou2023triplane}, which employs two transformer-based networks, a point decoder and a triplane decoder. The point decoder generates point clouds from a single image, acting as the explicit 3D prior for the triplane decoder to predict Gaussian features for each point. Another method, AGG~\cite{xu2024agg}, decomposes the generation of 3D Gaussian locations and appearance attributes by first generating a coarse 3D representation and subsequently upsampling it via a 3D Gaussian super-resolution module. It is a promising direction to improve the performance of 3D human modeling while incorporating better 3D representation like Mosaic-SDF~\cite{yariv2023mosaic} would further improve the quality.

\subsubsection{Real-time Gaussian-based Generation}\label{realtime}

Inspired by PIFu~\cite{saito2019pifu} that uses an image encoder and MLP for occupancy prediction based on pixel-aligned image features, GPS-Gaussian~\cite{zheng2023gps} proposes to learn a 2D Gaussian parameter map from sparse-view RGB images of human-centered scenes to reconstruct free-viewpoint renderings. Notably, GPS-Gaussian achieves real-time rendering of dynamic scenes by efficiently querying the 2D Gaussian parameter map. However, directly applying this method to a 3D foundation model or large-scale reconstruction model is challenging due to the complexity of training the corresponding encoder and decoder. Finding solutions to incorporate 2D Gaussian parameter maps into generative pipelines remains a valuable open problem, given its potential for real-time 3D generation.